\documentclass[11pt, letterpaper, logo, onecolumn, copyright]{main}

\usepackage[authoryear, sort&compress, round]{natbib}

\usepackage[inkscapeformat=png]{svg}

\usepackage[most, breakable, skins]{tcolorbox}

\tcbuselibrary{skins}
\usepackage{lipsum}
\usepackage{tabularx}
\usepackage{afterpage}
\usepackage{booktabs}
\usepackage{subcaption}
\usepackage{makecell}
\usepackage{multirow}
\usepackage{bm}
\usepackage{multicol}
\usepackage{array}
\usepackage{float}
\usepackage{listings, listings-rust}
\usepackage{fontawesome5}
\usepackage{hyperref}
\usepackage{amssymb,graphicx}
\usepackage[dvipsnames]{xcolor}
\usepackage{cleveref}
\usepackage{longtable}
\usepackage{pdflscape}
\usepackage{adjustbox}
\usepackage{nicematrix}
\usepackage{CJKutf8}
\usepackage{ragged2e}
\usepackage{colortbl}
\usepackage{enumitem}
\usepackage[ruled,linesnumbered]{algorithm2e}
\usepackage{pifont}
\usepackage{amsmath}
\usepackage{amsthm}  
\usepackage{mathrsfs} 
\usepackage{circledsteps}
\usepackage{diagbox}
\usepackage{bookmark}

\usepackage{wrapfig}
\usepackage{xspace}
\usepackage{tikz}
\usepackage[normalem]{ulem}

\usepackage{pgfplots}
\usepackage{pgfplotstable}
\pgfplotsset{compat=1.18}

\usepackage[textsize=tiny]{todonotes}
\usepackage{enumitem}
\usepackage{multirow}
\usepackage{colortbl}
\usepackage{pifont}
\usepackage{tabularx}
\usepackage{stfloats}
\usepackage{floatflt}

\definecolor{mymyblue}{RGB}{78,102,145}
\definecolor{mymyred}{RGB}{184,71,77}

\definecolor{medgray55}{gray}{0.55}
\definecolor{medgray}{gray}{0.7}
\definecolor{litegray}{gray}{0.9}
\definecolor{gblue}{RGB}{210, 227, 252}
\definecolor{gred}{RGB}{250, 210, 207}
\definecolor{gyellow}{RGB}{254, 239, 195}
\definecolor{ggreen}{RGB}{206, 234, 214}
\definecolor{gorange}{RGB}{254, 223, 200}

\definecolor{gblue9}{RGB}{23, 78, 166}
\definecolor{gred9}{RGB}{165, 14, 14}
\definecolor{gyellow9}{RGB}{227, 116, 0}
\definecolor{ggreen9}{RGB}{13, 101, 45}
\definecolor{gorange9}{RGB}{176, 96, 0}

\definecolor{myblue}{rgb}{0,0,1}
\definecolor{myred}{rgb}{1,0,0}
\definecolor{mylightgray}{gray}{0.95}
\definecolor{myCite}{HTML}{1C4587}

\definecolor{highlightblue}{HTML}{185ABC}
\definecolor{cellHighlight}{HTML}{dbefff}

\usepackage{minitoc}

\noptcrule


\newcolumntype{L}[1]{>{\raggedright\let\newline\\\arraybackslash\hspace{0pt}}m{#1}}
\newcolumntype{C}[1]{>{\centering}m{#1}}

\newcolumntype{R}[1]{>{\raggedleft\let\newline\\\arraybackslash\hspace{0pt}}m{#1}}

\definecolor{ao}{rgb}{0.0, 0.0, 1.0}

\newcommand\vcent[1]{\vcenter{\hbox{#1}}}

\newcommand\loudspeaker[1][3]{\ensuremath{\vcent{\rule{.6ex}{.6ex}}\kern-.5ex
  \vcent{\scalebox{.6}[1]{\rotatebox[origin=center]{90}{$\blacktriangle$}}}
  \ifnum#1>0\relax\kern.05ex\vcent{\scalebox{.4}{\ttfamily)}}
  \ifnum#1>1\relax\kern-.4ex\vcent{\scalebox{.56}{\ttfamily)}}
  \ifnum#1>2\relax\kern-.55ex\vcent{\scalebox{.7}{\ttfamily)}}
  \fi\fi\fi}
}

\makeatletter
\renewcommand\subparagraph{
 \@startsection {subparagraph}{5}{\z@ }{3.25ex \@plus 1ex
 \@minus .2ex}{-1em}{\normalfont \normalsize \bfseries }}
\makeatother

\let\cite\citep
\hypersetup{
  citecolor = myCite,  
  linkcolor = myCite,   
  urlcolor  = myCite
}

\title{Can RL Improve Generalization of LLM Agents? \\ An Empirical Study}
\author{
    Zhiheng Xi$^1$$^{* \dag}$,  Xin Guo$^1$$^*$, Jiaqi Liu$^1$, Jiazheng Zhang$^1$, Yutao Fan$^3$, Zhihao Zhang$^1$, \\ \textbf{Shichun Liu$^1$, Mingxu Chai$^1$, Xiaowei Shi$^2$, Yitao Zhai$^2$, Xunliang Cai$^2$, Tao Gui$^{1}$$^\dag$,} \\  \textbf{Qi Zhang$^1$$^\dag$, Xuanjing Huang$^1$$^\dag$} \\
$^1$Fudan University \ $^2$Meituan \ $^3$Shanghai Artificial Intelligence Laboratory\\
\texttt{zhxi22@m.fudan.edu.cn, \{tgui,qz,xjhuang\}@fudan.edu.cn} 
}

\begin{abstract}
Reinforcement fine-tuning (RFT) has shown promise for training LLM agents to perform multi-turn decision-making based on environment feedback. However, most existing evaluations remain largely in-domain—training and testing are conducted in the same environment or even on the same tasks.  In real-worlddeployment, agents may operate in unseen environments with different background knowledge, observation spaces, and action interfaces. To characterize the generalization profile of RFT under such shifts, we conduct a systematic study along three axes: (1) within-environment generalization across task difficulty, (2) cross-environment transfer to unseen environments, and (3) sequential multi-environment training to quantify transfer and forgetting. Our results show that RFT generalizes well across task difficulty within an environment, but exhibits weaker transfer to unseen environments, which correlates with shifts in both semantic priors and observation/action interfaces. In contrast, sequential training yields promising downstream gains with minimal upstream forgetting, and mixture training across environments improves the overall balance. We further provide detailed analyses and deeper insights, and hope our work helps the community develop and deploy generalizable LLM agents.
\end{abstract}

\begin{document}

\doparttoc
\faketableofcontents

\begingroup
  \renewcommand\thefootnote{}
  \footnote{\textsuperscript{*}Equal contribution.
            \textsuperscript{\dag}Corresponding authors.}
\endgroup

\maketitle

\section{Introduction}

Reinforcement fine-tuning (RFT) has emerged as a promising post-training paradigm for improving large language model (LLM) agents on complex interactive tasks such as web navigation and software engineering \citep{zhou2023webarena, deng2023mind2web, he2024webvoyager, jimenez2023swe, zan2025multi, merrill2026terminal}. In RFT, an agent is trained to make a sequence of intelligent decisions that maximize task-specific objectives based on environment feedback \citep{DBLP:journals/corr/abs-2509-08755,mai2025agent}. 
By optimizing for long-horizon outcomes, RFT is often associated with stronger agentic behaviors such as instruction following \citep{ouyang2022training, bai2022constitutional}, planning \citep{yao2022react, shinn2023reflexion}, reasoning \citep{lightman2023let, uesato2022solving}, and tool use \citep{li2025deepagent, li2023tool}.

Despite rapid progress, most empirical evidence for RFT focuses on in-domain evaluation, where training and testing are conducted within the same environment—and often even on closely overlapping tasks \citep{zhou2023webarena, liu2023agentbench}.
In real-world deployment, however, agents frequently encounter unseen environments that differ not only in task instances, but also in background knowledge, observation spaces, and action interfaces  \citep{ruan2023tptu, xi2025rise}. Such shifts can fundamentally reshape the interaction dynamics faced by agents, including which observations are informative, what actions are feasible, and how failures can be corrected.
This raises a practical and important research question: do the improvements brought by RFT generalize beyond the training distribution?

\begin{wrapfigure}{r}{0.55\textwidth}
  \begin{minipage}{\linewidth}
  \centering
  \vspace{-12pt}
  \includegraphics[width=\linewidth]{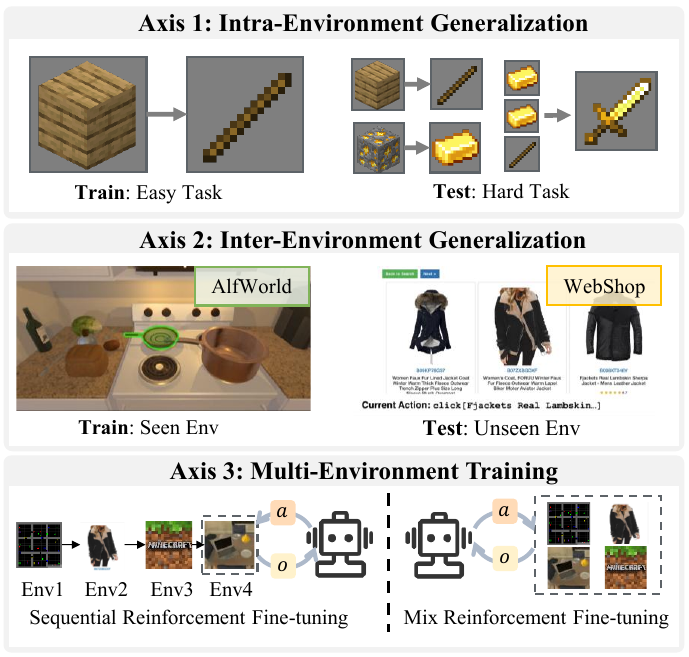}
  \captionof{figure}{
    An overview of three axes we study.
      }
  \label{fig:main}
  \vspace{10pt}
  \captionof{table}{Characteristics of each environment we study. The columns ``DEN.'', ``VAL.'', ``ACT.'', ``KNWL.'' and ``STR.'' indicate whether the rewards are \textit{dense}, action validation is \textit{strict}, \textit{valid action lists} are provided per step, \textit{world knowledge} is required, and observations are \textit{structured}, respectively. The environments vary along these dimensions.}
  \vspace{10pt}
  \label{tab:compare_environments}
      \resizebox{\linewidth}{!}{
        \begin{tabular}{ccccccc}
          \toprule
          \textbf{Environment} & \textbf{Types} & \textbf{DEN.} & \textbf{VAL.} & \textbf{ACT.} & \textbf{KNWL.} & \textbf{STR.} \\
          \midrule
          WebShop & \emph{Web} & \ding{52} & \ding{55} & \ding{55} & \ding{52} & \ding{52} \\
          SearchQA & \emph{Search} & \ding{55} & \ding{52} & \ding{55} & \ding{52} & \ding{52} \\
          TextCraft & \emph{Game} & \ding{55} & \ding{52} & \ding{55} & \ding{55} & \ding{55} \\
          AlfWorld & \emph{Household} & \ding{55} & \ding{52} & \ding{55} & \ding{55} & \ding{55} \\
          BabyAI & \emph{Embodied} & \ding{52} & \ding{55} & \ding{52} & \ding{55} & \ding{55} \\
          \bottomrule
        \end{tabular}
        }
    \end{minipage}
\end{wrapfigure}

To bridge this gap, we conduct a systematic study to explore how RFT impacts the generalization and transferability of LLM agents across three axes (Figure \ref{fig:main}).
First, we investigate task-level generalization under the same environment by reviewing varying difficulty performance when RFT on a specific subset of tasks (Section \ref{sec:study1}); 
Second, we examine environment-level generalization to assess whether RFT maintains efficacy in unseen environments, where agents must navigate shifts in background knowledge alongside changes in observation and action spaces (Section \ref{sec:study2}).  
Third, we analyze sequential training across environments to characterize the dynamics of transfer and forgetting, comparing this approach against joint training on environmental mixtures (Section \ref{sec:study3}). 
Together, these three axes provide a comprehensive, systematic framework for understanding RFT-driven generalization and transfer in LLM agents.

Our analysis covers five agentic environments that vary in background and core properties, as shown in Table \ref{tab:compare_environments} and Table \ref{tab:action_space}. 
We highlight several key findings:
(1)~Intra-environment generalization, within a consistent environment, RFT-trained agents exhibit significant generalization, and easy-to-hard curriculum learning further boosts performance gains.
(2)~Inter-environment sensitivity, while RFT enhances agentic capabilities, its generality to unseen environments exhibits fluctuations. Our analysis reveals that this across-environment generalization is sensitive to the required prior knowledge, observation spaces, and action spaces.
(3) Multi-environment training. Employing multi-environment training strategies substantially enhances generalization: sequential RFT enables effective transfer to new environments while maintaining performance on upstream ones, and mix RFT across multiple environments achieves consistently strong results overall.
We further provide insights that shed light on the mechanisms underlying these behaviors. 
Overall, we hope these findings inform the development of more generalizable LLM agents and their practical deployment in real-world settings.

\section{Related Work}

\subsection{RL for LLM Agents} Reinforcement learning (RL) has been widely adopted in LLM training, with paradigms such as reinforcement learning from human feedback (RLHF) \citep{DBLP:conf/nips/ChristianoLBMLA17, DBLP:conf/nips/Ouyang0JAWMZASR22, DBLP:journals/corr/abs-2212-08073} and reinforcement learning from verifiable rewards (RLVR) \citep{DBLP:journals/corr/SchulmanWDRK17, DBLP:conf/nips/RafailovSMMEF23, DBLP:journals/corr/abs-2501-12948} demonstrating strong effectiveness in improving instruction following, reasoning accuracy, and behavioral alignment \citep{DBLP:conf/nips/MuHHAVKLBSW24, guo2025ifdecorator, DBLP:journals/corr/abs-2504-21801}. Building upon these advances, recent work has extended reinforcement learning to LLM agents to enhance multi-step decision making \citep{zhai2024fine, zhai2025enhancing, wang2025offline}, long-horizon planning \citep{DBLP:journals/corr/abs-2403-02502, DBLP:journals/corr/abs-2509-08755, DBLP:journals/corr/abs-2505-20732}, and interaction with external environments \citep{tan2024cradle, DBLP:conf/nips/FengHHLZZL24, DBLP:journals/corr/abs-2502-01600}. Across diverse agent settings, RL has been shown to strengthen agents’ abilities to decompose complex tasks, coordinate reasoning with action~, and adapt policies through environment feedback \citep{tian2024reinforcement}, enabling more effective information gathering \citep{DBLP:conf/iclr/ZhangKC25, ramrakhya2025grounding, DBLP:journals/corr/abs-2503-09516}, iterative self-correction \citep{DBLP:conf/iclr/KumarZASCSBIBRZ25, DBLP:journals/corr/abs-2502-12853, zeng2025evolving}, and tool utilization \citep{DBLP:journals/corr/abs-2511-01934, DBLP:journals/corr/abs-2504-11536, DBLP:journals/corr/abs-2510-21618}. Despite these successes, most RL-based agent methods are evaluated primarily under in-domain settings with similar task distributions and interfaces, leaving open the question of how well such learned decision-making policies generalize across tasks or transfer to unseen environments.

\subsection{Generalization and Forgetting by Post-training}

Recent work has examined how post-training strategies affect the generalization and forgetting behaviors of LLMs \citep{DBLP:journals/tist/ZhaoCYLDCWYD24, zhang2025reinforcement, DBLP:journals/corr/abs-2510-23081}. Supervised fine-tuning (SFT) is effective for improving in-distribution performance but often leads to over-specialization and degradation of general capabilities due to representation and distributional drift \citep{DBLP:conf/iclr/KumarRJ0L22, DBLP:journals/corr/abs-2308-08747, DBLP:conf/iclr/KothaSR24, wang2024generalization}. In contrast, RL, particularly RLVR, tends to better preserve pre-trained representations by optimizing trajectory-level objectives rather than directly fitting target distributions, enabling more robust transfer across tasks and domains \citep{DBLP:journals/corr/abs-2507-00432, DBLP:journals/corr/abs-2510-18874, cheng2025revisiting, chu2025sft}. However, RL can also induce negative interference and winner-take-all dynamics, reducing behavioral coverage and leading to reasoning boundary shrinkage \citep{DBLP:journals/corr/abs-2510-02230, DBLP:journals/corr/abs-2506-19733, sun2025omega}. Notably, most existing studies focus on static and single-turn LLM tasks, whereas we extend the study of post-training generalization and forgetting to multi-turn LLM agents and systematically evaluate these effects across different environments with distinct observation and action spaces.

\section{Preliminaries}

\subsection{Preliminaries}

\paragraph{Task Formulation.}

We follow the ReAct interaction paradigm \citep{yao2022react} to formulate a multi-turn decision-making task. The task can be represented by the tuple $(\mathcal{U}, \mathcal{S}, \mathcal{A}, \mathcal{O}, \mathcal{T}, \mathcal{R})$. Here, $\mathcal{U}$ denotes the instruction space, $\mathcal{S}$ is the state space, $\mathcal{A}$ is the action space, and $\mathcal{O}$ represents the observation space. The function $\mathcal{T}: \mathcal{S} \times \mathcal{A} \rightarrow \mathcal{S}$ denotes the deterministic state transition function, and $\mathcal{R}: \mathcal{U} \times \mathcal{S} \rightarrow \mathbb{R}$ is the reward function.

Given a task instruction $u \in \mathcal{U}$, the agent operates based on a policy $\pi_{\theta}$ (an LLM parameterized by $\theta$). In each interaction step $t$, the state $s_t$ consists all previous dialogues and their resulting observations, \textit{i.e.}, $s_t = (a_0, o_0, \dots, a_{t-1}, o_{t-1})$. The agent generates an action $a_t \sim \pi_\theta(\,\cdot \mid u, s_t)$. The action $a_t$ includes an internal reasoning trace and a interaction to the environment. The agent then receives an observation $o_k \in \mathcal{O}$ from the environment. This interaction loop continues until the tesk is completed or the maximum number of turns is reached, resulting in a complete trajectory:
\begin{equation}
    \tau = \{u, (a_0, o_0), (a_1, o_1), \dots, (a_T, o_{T})\}
\end{equation}
The environment provides a terminal reward $\mathcal{R}(\tau) \in [0,1]$ upon task completion.

\paragraph{Policy Gradient.}

In RL, our objective is to optimize the policy parameters $\theta$ to maximize the expected cumulative reward over all possible trajectories for the given task \citep{sutton1998reinforcement}:
\begin{equation}
    \max_{\theta} J(\theta) = \mathbb{E}_{\tau \sim \pi_{\theta}}[\mathcal{R}(\tau)]
\end{equation}

To optimize the objective function $J(\theta)$, we utilize policy gradient methods \citep{sutton1999policy}. Unlike value-based methods that approximate a value function, policy gradient methods directly search the policy parameter space. The core idea is to perform gradient ascent to update $\theta$, increasing the probability of trajectories that yield high rewards. The vanilla policy gradient is formulated as:
\begin{equation}
    \nabla_{\theta} J(\theta) = \mathbb{E}_{\tau \sim \pi_{\theta}} \left[ \mathcal{R}(\tau) \sum_{t=0}^{T} \nabla_{\theta} \log \pi_{\theta}(a_t|s_t) \right]
\end{equation}
Based on this gradient estimation, the parameters are updated via $\theta_{new} = \theta_{old} + \alpha \nabla_{\theta} J(\theta)$, where $\alpha$ is the learning rate. However, standard policy gradient methods often suffer from high variance, leading to unstable training dynamics. To address this challenge, we adopt the widely-used GRPO algorithm \citep{DBLP:journals/corr/abs-2501-12948}, detailed in Appendix \ref{appendix:grpo}.

\subsection{Experimental Settings}\label{sec:experiment_setting}

\paragraph{Environment Setup.}
We select five representative agent environments, including WebShop \cite{yao2022webshop}, SearchQA \cite{dunn2017searchqa}, TextCraft \cite{sanghi2022textcraft}, AlfWorld \cite{shridhar2020alfworld}, and BabyAI \cite{chevalier2018babyai}, with characteristics of each environment summarized in Table \ref{tab:compare_environments}. Among them, WebShop is an interactive web shopping environment; SearchQA is a Q\&A environment augmented with context from a search engine; ALFWorld requires agents to explore rooms and execute tasks in a household setting; BabyAI is an interactive grid world simulator; and TextCraft is a game environment for crafting items in Minecraft. 
The task data $\mathcal{U}$ is sourced from AgentGym \cite{xi2025agentgym}, with detailed statistics and action spaces for each environment provided in Appendix~\ref{appendix:environments}.

\paragraph{Training Setup.}
We perform RFT training using the AgentGym-RL framework\footnote[1]{\url{https://github.com/woooodyy/AgentGym-RL}} \cite{DBLP:journals/corr/abs-2509-08755} with Qwen2.5-3B-Instruct and Qwen2.5-7B-Instruct \cite{qwen2.5}. Following the ReAct \cite{yao2022react} paradigm, each action receives real-time environment feedback. For all experiments, we sample 8 trajectories per task and set the maximum response length to 8192 tokens. Following \citet{xi2025agentgym}, we set the maximum interaction turns $K$ for each environment as follows: 10 for WebShop, 5 for SearchQA, 30 for AlfWorld, 10 for BabyAI, and 15 for TextCraft.

\paragraph{Evaluation Setup.}
As summarized in Table~\ref{tab:compare_environments}, WebShop and BabyAI provide dense rewards, while SearchQA, AlfWorld, and TextCraft use binary rewards. We follow \citet{DBLP:journals/corr/abs-2501-12948} and adopt exact matching, counting an action as correct only if it exactly matches the reference. To reduce randomness, we report \texttt{avg@8} as the main metric, and also measure the average interaction turns ($\bar{k}$) and generated tokens ($\bar{l}$) for efficiency. For fair comparison, all agents are evaluated with the maximum number of interaction turns set to $K=20$ during testing.

\section{Does RFT Yield Generalization Across Task Difficulties within the Same Environment?}\label{sec:study1}

Reinforcement fine-tuning trains LLM agents through interaction with the environment, allowing them to learn its dynamics and adapt over time \citep{song2024trial, cui2025entropy}.
Yet even with a fixed action and observation space, tasks within the same environment can differ substantially in exploration depth and information accessibility. 
In this section, we investigate \emph{whether RFT policies learned on a subset of tasks transfer to other tasks of differing difficulty within the same environment}.

\begin{table*}[t]
  \caption{Results of generalization across task difficulties within the same environment.}
  \vspace{-10pt}
  \label{tab:study1_1}
  \begin{center}
      \resizebox{\textwidth}{!}{
        \begin{tabular}{lrrrrrrrrrrrrrrr}
          \toprule
          \multirow{2}{*}{\textbf{Models}} & \multicolumn{3}{c}{\textbf{WebShop}} & \multicolumn{3}{c}{\textbf{SearchQA}} & 
          \multicolumn{3}{c}{\textbf{TextCraft}} & \multicolumn{3}{c}{\textbf{AlfWorld}} & \multicolumn{3}{c}{\textbf{BabyAI}} \\
          & \emph{easy} & \emph{hard} & \emph{all} & \emph{easy} & \emph{hard} & \emph{all} & \emph{easy} & \emph{hard} & \emph{all} & \emph{easy} & \emph{hard} & \emph{all} & \emph{easy} & \emph{hard} & \emph{all} \\
          \midrule
          \multicolumn{16}{c}{\emph{\textbf{Qwen2.5-3B-Instruct}}} \\
          base model & $21.7$ & $10.6$ & \cellcolor{gray!10}$15.3$ & $63.7$ & $6.5$ & \cellcolor{gray!10}$23.7$ & $23.7$ & $2.3$ & \cellcolor{gray!10}$14.5$ & $26.8$ & $8.0$ & \cellcolor{gray!10}$13.2$ & $71.5$ & $48.0$ & \cellcolor{gray!10}$61.6$ \\
          train w/ $\mathcal{U}_\text{easy}$ & $90.3$ & $75.3$ & \cellcolor{gray!10}$81.6$ & $82.7$ & $16.9$ & \cellcolor{gray!10}$36.6$ & $97.6$ & $49.4$ & \cellcolor{gray!10}$76.9$ & $93.2$ & $82.0$ & \cellcolor{gray!10}$85.1$ & $93.4$ & $79.0$ & \cellcolor{gray!10}$87.3$ \\
          train w/ $\mathcal{U}_\text{hard}$ & $86.1$ & $84.5$ & \cellcolor{gray!10}$85.2$ & $72.5$ & $19.4$ & \cellcolor{gray!10}$35.3$ & $95.0$ & $42.2$ & \cellcolor{gray!10}$72.3$ & $96.1$ & $92.9$ & \cellcolor{gray!10}$93.8$ & $92.3$ & $77.3$ & \cellcolor{gray!10}$86.0$ \\
          train w/ $\mathcal{U}$ & $92.8$ & $84.3$ & \cellcolor{gray!10}$87.9$ & $87.6$ & $22.1$ & \cellcolor{gray!10}$41.8$ & $93.9$ & $47.4$ & \cellcolor{gray!10}$73.9$ & $97.0$ & $89.8$ & \cellcolor{gray!10}$91.8$ & $93.2$ & $77.5$ & \cellcolor{gray!10}$86.6$ \\
          $\mathcal{U}_\text{easy}+\mathcal{U}_\text{hard}$ & $93.6$ & $85.2$ & \cellcolor{gray!10}$88.7$ & $82.1$ & $19.8$ & \cellcolor{gray!10}$38.5$ & $93.9$ & $52.9$ & \cellcolor{gray!10}$76.3$ & $97.3$ & $93.4$ & \cellcolor{gray!10}$94.4$ & $94.5$ & $82.2$ & \cellcolor{gray!10}$89.3$ \\
          $\mathcal{U}_\text{hard}+\mathcal{U}_\text{easy}$ & $90.2$ & $82.4$ & \cellcolor{gray!10}$85.7$ & $82.0$ & $18.7$ & \cellcolor{gray!10}$37.7$ & $97.8$ & $40.1$ & \cellcolor{gray!10}$73.0$ & $95.5$ & $92.3$ & \cellcolor{gray!10}$93.6$ & $94.5$ & $80.4$ & \cellcolor{gray!10}$88.6$ \\
          \midrule
          \multicolumn{16}{c}{\emph{\textbf{Qwen2.5-7B-Instruct}}} \\
          base model & $44.1$ & $17.4$ & \cellcolor{gray!10}$28.6$ & $79.6$ & $10.4$ & \cellcolor{gray!10}$31.2$ & $46.9$ & $16.0$ & \cellcolor{gray!10}$33.6$ & $40.2$ & $21.4$ & \cellcolor{gray!10}$26.6$ & $80.1$ & $47.9$ & \cellcolor{gray!10}$67.0$ \\
          train w/ $\mathcal{U}_\text{easy}$ & $88.1$ & $77.5$ & \cellcolor{gray!10}$82.0$ & $85.8$ & $21.1$ & \cellcolor{gray!10}$40.5$ & $98.2$ & $47.4$ & \cellcolor{gray!10}$76.4$ & $95.2$ & $89.7$ & \cellcolor{gray!10}$91.2$ & $95.2$ & $76.1$ & \cellcolor{gray!10}$87.2$ \\
          train w/ $\mathcal{U}_\text{hard}$ & $87.9$ & $81.4$ & \cellcolor{gray!10}$84.2$ & $93.3$ & $27.0$ & \cellcolor{gray!10}$46.9$ & $99.3$ & $57.3$ & \cellcolor{gray!10}$81.3$ & $97.5$ & $94.6$ & \cellcolor{gray!10}$95.4$ & $92.8$ & $77.9$
          & \cellcolor{gray!10}$86.5$ \\
          train w/ $\mathcal{U}$ & $92.9$ & $81.9$ & \cellcolor{gray!10}$86.5$ & $91.8$ & $26.6$ & \cellcolor{gray!10}$46.1$ & $95.2$ & $52.3$ & \cellcolor{gray!10}$80.9$ & $97.7$ & $97.2$ & \cellcolor{gray!10}$92.0$ & $95.6$ & $74.5$ & \cellcolor{gray!10}$88.8$ \\
          $\mathcal{U}_\text{easy}+\mathcal{U}_\text{hard}$ & $90.6$ & $77.5$ & \cellcolor{gray!10}$83.0$ & $89.8$ & $25.5$ & \cellcolor{gray!10}$44.8$ & $98.7$ & $64.0$ & \cellcolor{gray!10}$83.8$ & $96.1$ & $92.6$ & \cellcolor{gray!10}$93.6$ & $95.5$ & $82.9$ & \cellcolor{gray!10}$90.2$ \\
          $\mathcal{U}_\text{hard}+\mathcal{U}_\text{easy}$ & $89.1$ & $79.1$ & \cellcolor{gray!10}$83.3$ & $87.6$ & $21.5$ & \cellcolor{gray!10}$42.3$ & $99.8$ & $57.0$ & \cellcolor{gray!10}$81.4$ & $97.7$ & $88.5$ & \cellcolor{gray!10}$91.1$ & $92.5$ & $78.2$ & \cellcolor{gray!10}$86.5$ \\
          \bottomrule
        \end{tabular}
    }
  \end{center}
\end{table*}

\paragraph{Setting.}

Following practice in previous work \citep{bengio2009curriculum, mukherjee2023orca}, we categorize the tasks $\mathcal{U}$ into \emph{easy} and \emph{hard} difficulty levels (denoted as $\mathcal{U}_\text{easy}$ and $\mathcal{U}_\text{hard}$) based on the \texttt{avg@8} results of the Qwen2.5-7B-Instruct model, while ensuring a balanced distribution of data between the two difficulty levels. The same categorization is applied to the test set. Detailed data statistics for each environment are provided in Appendix~\ref{appendix:environments}.

\paragraph{RFT demonstrates strong tranferability across varying difficulty levels within the same environment.} 
As shown in Table \ref{tab:study1_1}, RFT demonstrates strong robustness when trained on data of varying difficulty levels within the same environment. Notably, with 7B model on WebShop, training on $\mathcal{U}_{\text{easy}}$ improves performance on the \textit{hard} testset by $60.1$ points, suggesting that RFT encourages agents to adapt to the environment, thereby enabling performance transfer across tasks with different difficulty levels and varying steps \citep{DBLP:journals/corr/abs-2507-00432, DBLP:journals/corr/abs-2510-18874}. 
Overall, we also find that training on $\mathcal{U}_\text{hard}$ yields more gains on test set, e.g., for 3B model in AlfWorld, training on $\mathcal{U}_\text{hard}$ outperforms training on $\mathcal{U}_\text{easy}$ by $8.7$ points. We attribute this trend to the richer failure signals and longer-horizon exploration induced by harder tasks, which push the RFT optimizer to seek better policies \citep{xu2023wizardlm, wang2023voyager}.

\paragraph{Curriculum learning can further enhance performance.} 
Additionally, we investigate the impact of the training sequence using data of varying difficulty levels. Results in Table \ref{tab:study1_1} indicate that training on $\mathcal{U}$ (i.e., mixture of $\mathcal{U}_\text{easy}$ and $\mathcal{U}_\text{hard}$) generally does not yield optimal performance. In most cases, training on $\mathcal{U}_\text{easy}$ first, followed by $\mathcal{U}_\text{hard}$, achieves the best results. Compared with using a single difficulty level, this easy-to-hard curriculum enables further performance improvement. For example, on BabyAI, training on $\mathcal{U}_\text{easy}+\mathcal{U}_\text{hard}$ outperforms training on $\mathcal{U}_\text{easy}$ and on $\mathcal{U}_\text{hard}$ by $2.0$ and $3.3$ points, respectively. This result validates the rationale and effectiveness of curriculum learning \cite{qi2024webrl, zhang2025learning}. 

\begin{wrapfigure}{r}{0.55\linewidth}
  \centering
  \vspace{-12pt}
  \includegraphics[width=0.99\linewidth]{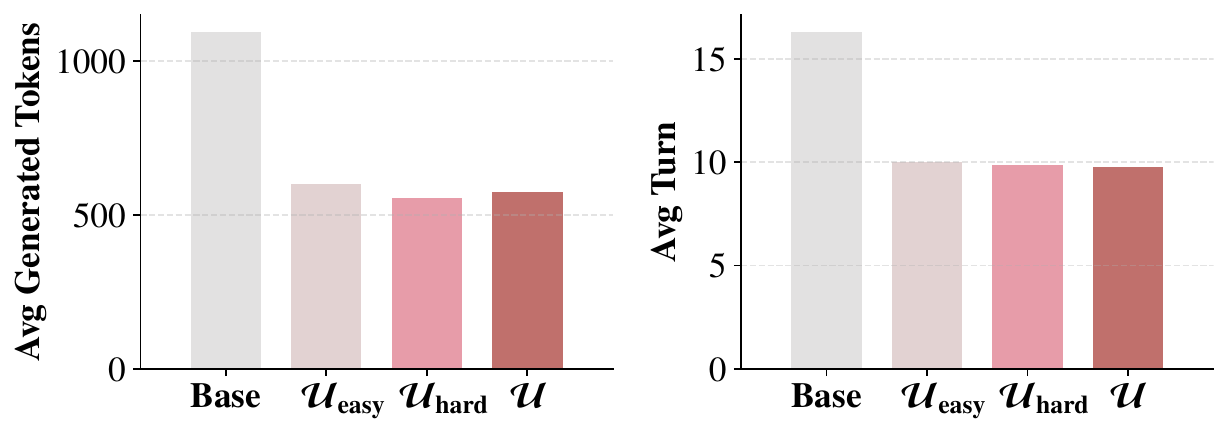}
  \caption{Average generated tokens and interactive turn of all environments for Qwen2.5-3B-Instruct model trained with varying-difficulty tasks.}
  \label{fig:study1}
\end{wrapfigure}

\paragraph{RFT training encourages agents to perform efficient exploration.}
Figure~\ref{fig:study1} reports the average number of interaction turns and generated tokens across different environments, with a more detailed breakdown provided in Figure~\ref{fig:avg_round_different_env} of Appendix~\ref{appendix:avg_turn_and_token}. We find that RFT enables agents to interact and explore more efficiently within the same environment. For example, when training the 3B model on $\mathcal{U}_{\text{easy}}$ in BabyAI, the average interaction turns decrease from $10.76$ to $4.19$, and the average trajectory length is reduced from $624.58$ to $160.60$ tokens. These results indicate that RFT not only improves success rates but also encourages more concise and goal-directed exploration, substantially enhancing interaction efficiency \citep{nakano2021webgpt, song2024trial}. A case can be found in Appendix \ref{appendix:case_study}.

\section{Does RFT Yield Generalization Across Different Environments?}\label{sec:study2}

In real-world scenarios, agents may encounter previously unseen environments and tasks, which is essential for practical deployment. We therefore investigate \emph{whether RFT improves performance in unseen environments with different background knowledge, observation space and action spaces}. Concretely, we perform RFT training within a single environment and then evaluate the trained agent on other environments to measure cross-environment generalization.

\begin{table*}[t]
  \caption{Results of generalization across different environments. Darker \textcolor{mymyred}{red} indicates greater performance improvement, while darker \textcolor{mymyblue}{blue} signifies more pronounced performance decline. The best result is in \textbf{bold}, while the second-best is marked with \underline{underline}.}
  \vspace{-10pt}
  \label{tab:study2_1}
  \begin{center}
  \resizebox{\textwidth}{!}{
        \begin{tabular}{lcccccccc}
          \toprule
          \textbf{Models} & \textbf{WebShop} & \textbf{SearchQA} & \textbf{TextCraft} & \textbf{AlfWorld} & \textbf{BabyAI} & $\mathbf{\Delta}$\textbf{Held-In} & $\mathbf{\Delta}$\textbf{Held-Out} & $\mathbf{\Delta}$\textbf{Overall}\\
          \midrule
          \multicolumn{9}{c}{\emph{\textbf{Qwen2.5-3B-Instruct}}} \\
          base model & $15.30$ & $23.66$ & $14.50$ & $13.19$ & $61.83$ & $-$ & $-$ & $-$ \\
          train w/ WebShop & \cellcolor{red!25}$87.86$ & \cellcolor{red!3}$25.84$ & \cellcolor{red!5}$18.50$ & \cellcolor{red!10}$23.06$ & \cellcolor{red!9}$70.91$ & \underline{$+72.56$} & $\mathbf{+6.29}$ & $\underline{+19.54}$ \\
          train w/ SearchQA & \cellcolor{red!8}$22.97$ & \cellcolor{red!25}$41.78$ & \cellcolor{red!9}$22.75$ & \cellcolor{blue!1}$12.13$ & \cellcolor{red!4}$65.72$ & $+18.13$ & $+4.69$ & $+\ \ 4.56$ \\
          train w/ TextCraft & \cellcolor{blue!2}$14.46$ & \cellcolor{blue!3}$22.16$ & \cellcolor{red!25}$73.88$ & \cellcolor{blue!3}$11.00$ & \cellcolor{red!2}$63.73$ & $+59.38$ & $-0.66$ & $+11.35$\\
          train w/ AlfWorld & \cellcolor{red!7}$21.86$ & \cellcolor{blue!1}$23.50$ & \cellcolor{red!10}$24.88$ & \cellcolor{red!25}$91.81$ & \cellcolor{red!3}$64.68$ & $\mathbf{+78.62}$ & $\underline{+4.91}$ & $\mathbf{+19.65}$ \\
          train w/ BabyAI & \cellcolor{red!12}$28.36$ & \cellcolor{blue!1}$22.63$ & \cellcolor{red!2}$16.75$ & \cellcolor{blue!9}$\ \ 4.50$ & \cellcolor{red!25}$86.55$ & $+24.72$ & $+1.40$ & $+\ \ 6.06$ \\
          \midrule
          \multicolumn{9}{c}{\emph{\textbf{Qwen2.5-7B-Instruct}}} \\
          base model & $28.59$ & $31.19$ & $33.63$ & $26.56$ & $67.00$ & $-$ & $-$ & $-$ \\
          train w/ WebShop & \cellcolor{red!25}$86.50$ & \cellcolor{red!2}$33.28$ & \cellcolor{red!7}$40.75$ & \cellcolor{blue!2}$24.13$ & \cellcolor{red!12}$79.21$ & \underline{$+57.91$} & $+4.75$ & \underline{$+15.38$} \\
          train w/ SearchQA & \cellcolor{red!15}$47.07$ & \cellcolor{red!25}$46.12$ & \cellcolor{red!2}$35.25$ & \cellcolor{blue!8}$16.75$ & \cellcolor{red!13}$80.33$ & $+14.93$ & \underline{$+5.91$} & $+\ \ 7.71$ \\
          train w/ TextCraft & \cellcolor{red!10}$38.30$ & \cellcolor{red!1}$32.19$ & \cellcolor{red!25}$80.88$ & \cellcolor{red!5}$31.50$ & \cellcolor{red!10}$77.95$ & $+47.25$ & $\mathbf{+6.65}$ & $+14.77$ \\
          train w/ AlfWorld & \cellcolor{red!6}$34.31$ & \cellcolor{blue!2}$29.59$ & \cellcolor{red!3}$36.13$ & \cellcolor{red!25}$92.00$ & \cellcolor{red!5}$72.91$ & $\mathbf{+65.44}$ & $+3.13$ & $\mathbf{+15.59}$ \\
          train w/ BabyAI & \cellcolor{blue!10}$10.25$ & \cellcolor{blue!2}$29.41$ & \cellcolor{red!5}$39.25$ & \cellcolor{red!2}$28.13$ & \cellcolor{red!25}$88.79$ & $+21.79$ & $-3.23$ & $+\ \ 1.77$ \\
          \bottomrule
        \end{tabular}
    }
  \end{center}
\end{table*}

\paragraph{Setting.}
We employ the base model (i.e., Qwen2.5-3B-Instruct or Qwen2.5-7B-Instruct) as the baseline and evaluate its \texttt{avg@8} metric across varying environments. 
To make these comparisons explicit, we define three metrics: $\Delta$Held-In, $\Delta$Held-Out, and $\Delta$Overall. $\Delta$Held-In measures the average improvement over the baseline when the training and test environments coincide. $\Delta$Held-Out measures the average improvement when the agent is evaluated on environments different from those seen during training. Finally, $\Delta$Overall reports the average improvement over the baseline across all evaluation settings.

\paragraph{Performance differs significantly between held-in and held-out environments.} 

Table \ref{tab:study2_1} reveals that agents exhibit generalization across different environments, yet a significant performance gap exists between held-in and held-out settings. Specifically, substantial gains are achieved under held-in conditions. In AlfWorld, performance improves by $78.62$ and $65.44$ points for the 3B and 7B models, respectively. In contrast, in held-out conditions, generalization remains possible in most cases, but with more modest gains. On average, the 3B and 7B models yield improvements of $3.32$ and $3.44$ points on unseen environments, respectively. 

\paragraph{In unseen environments, positive transfer is observed in most cases.}
Compared to the baseline, most trained agents can demonstrate transferability to unseen environments, even when these environments demand different background knowledge and operate under different action spaces. Notably, agents trained on WebShop, AlfWorld, and SearchQA consistently exhibit positive $\Delta$held-out. When evaluated on WebShop, agents trained on SearchQA yield performance gains of $7.67$ and $18.48$ points for the 3B and 7B models, respectively. This can be attributed to the similarity between WebShop and SearchQA as search-based environments. Specifically, the agent trained on SearchQA learns to formulate more flexible search queries, as well as efficient information extraction from complex results. An illustrative case is presented in Figure \ref{fig:case_webshop} in Appendix \ref{appendix:case_study} and discussed in Section \ref{sec:discussion}.

\paragraph{In unseen environments, negative effects may sometimes occur.}
However, agents trained on TextCraft and BabyAI may struggle to generalize to other environments, e.g., after training on BabyAI, the 7B models show average negative improvements of $-3.23$ on held-out tasks, and even drop sharply from $28.59$ to $10.25$ on WebShop. Through careful analysis, we find that because the BabyAI environment provides available actions at each step, the agent gradually becomes dependent on this information during training, leading to a decline in long-horizon reasoning capability. When faced with other environments, it fails to accurately use the valid action, resulting in a sharp performance drop. A case of interaction between the agent and BabyAI environment is provided in Appendix \ref{appendix:case_study}.

\begin{sloppypar}
\paragraph{Cross-environment generalization performance varies significantly across target environments.}
Moreover, our observations reveal that cross-environment generalization performance is highly dependent on the target environment. 
On the one hand, agents trained on nearly all source environments generalize effectively to TextCraft and BabyAI, as these domains rely less on specific background knowledge, thereby allowing acquired skills to transfer more readily. On the other hand, effective transfer to AlfWorld and SearchQA proves to be significantly more challenging. We attribute this to the strict action validation and sparse feedback inherent in these two environments. For instance, AlfWorld responds uniformly with ``Nothing happens.'' to all invalid actions, offering no instructional guidance for improvement.
\end{sloppypar}

\section{How Does Sequential RFT Across Environments Affect Transfer and Forgetting?}\label{sec:study3}

Beyond zero-shot transfer to unseen environments without additional training, we further investigate the effect of sequential training across multiple environments on agentic behaviors, specifically the resulting learning dynamics on memorizing downstream tasks or forgetting on upstream ones.
In our experimental setup, a model initially converged in one environment is further trained on a second environment, after which we assess both its retention of performance on the original environment and its adaptation to the new one. Finally, we extend this analysis to sequential training across five environments, comparing this strategy with joint training across a mixture of environments.

\paragraph{Setting.}
We conduct 20 two-stage experiments by sequentially pairing each of the five environments as the upstream and the downstream, as well as several five-stage training experiments. Using agents trained on single-environment as baselines, we evaluate anti-forgetting by upstream performance retention and transferability by downstream performance gains. Furthermore, we compare the five-stage sequential approach with joint training, in which the agent is trained on all available data from every environment in a randomly mixed manner.

\begin{figure*}[t]
  \centering
  \includegraphics[width=\textwidth]{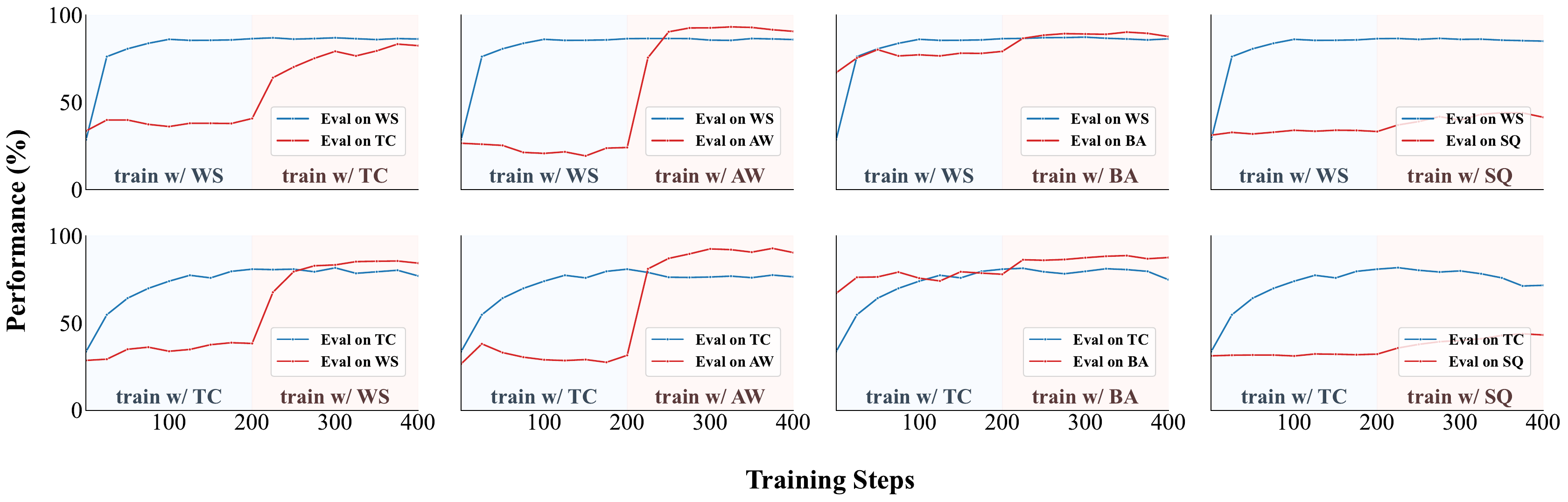}
  \caption{Training dynamics of forgetting and transfer in sequential two-stage cross-environment training with Qwen2.5-7B-Instruct, where \textcolor{mymyblue}{blue} and \textcolor{mymyred}{red} denote the upstream environment and the downstream environment, respectively.}
  \label{fig:study3_dynamics}
\end{figure*}

\paragraph{Sequential training demonstrates consistent anti-forgetting and transferability.}
Figure \ref{fig:study3_dynamics} illustrates the performance dynamics of 8 two-stage sequential training scenarios, with others and the final results summarized in Figure \ref{fig:study3_dynamics_others} and Table \ref{tab:study3} in Appendix \ref{appendix:study3}. Overall, sequential training matches or exceeds single-task performance on the downstream environment while largely preserving upstream performance. For instance, when the WebShop-pre-trained agent is further trained on TextCraft, it boosts performance on TextCraft from the single-task baseline of $80.88$ to $82.50$. Meanwhile, performance on WebShop experiences only a minor fluctuation, shifting from $86.5$ to $86.32$. These findings suggest that sequential training endows agents with stable capabilities for transfer and resistance to forgetting.

\begin{figure*}[t]
  \centering
  \includegraphics[width=\textwidth]{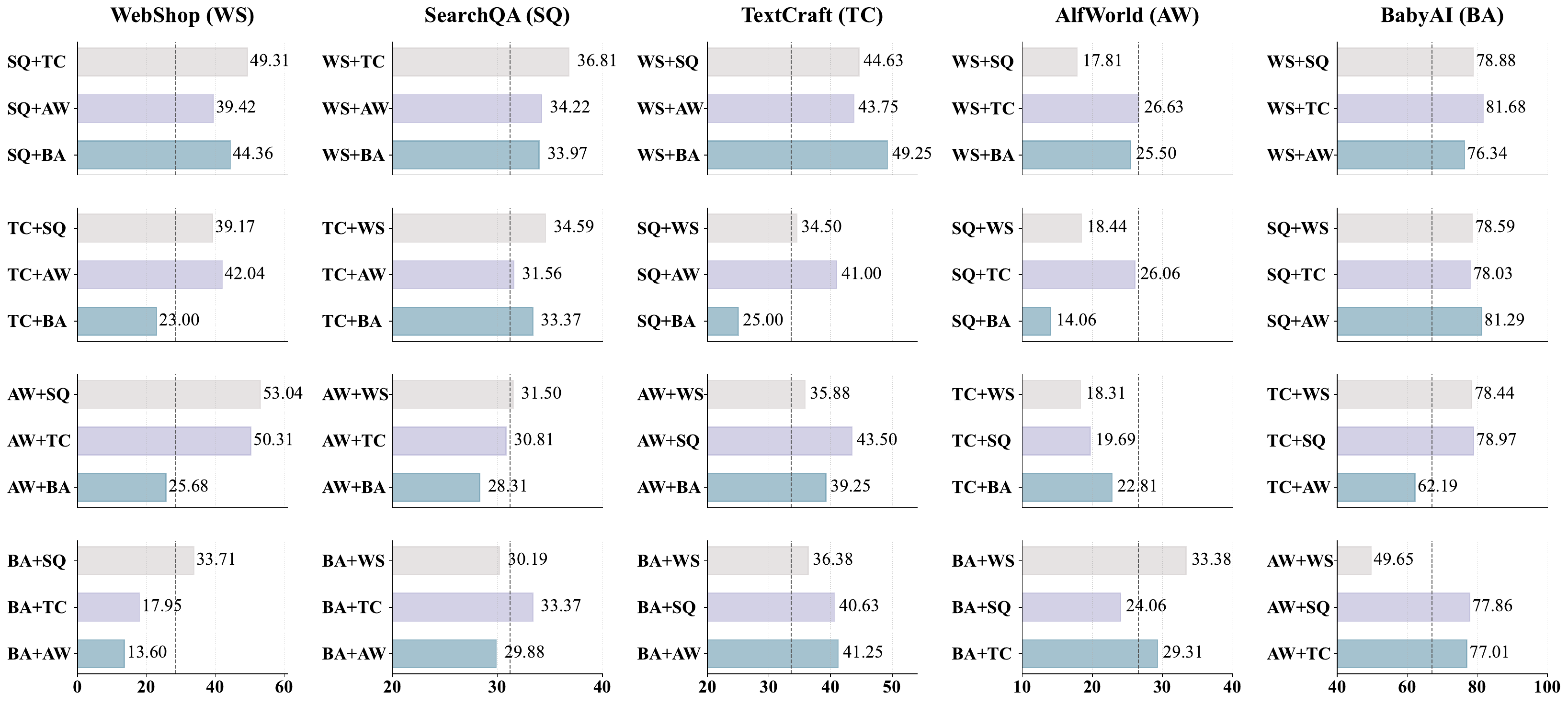}
  \caption{Generalization of sequential two-stage cross-environment training with Qwen2.5-7B-Instruct. Across five environments (WS, SQ, TC, AW, BA), the figure presents the generalization performance on the \emph{three unseen environments} following sequential training on two environments. Each subplot corresponds to a fixed first environment, with \emph{dashed line} indicating the baseline performance.}
  \label{fig:study3_generalization}
\end{figure*}

\paragraph{Generalization in multi-environment training is highly correlated with single-environment training.} 
Figure~\ref{fig:study3_generalization} demonstrates the generalization performance of multi-environment training, revealing a strong alignment with the generalization patterns in single-environment training. First, environments that yield poor generalization in single-environment settings continue to be detrimental. For instance, the agent trained on BabyAI exhibits severe negative effect on WebShop. Even when BabyAI is trained sequentially after other generalizable environments (i.e., TC+BA, AW+BA), it drastically degrades their WebShop performance. Notably, the score of TC+BA dropped sharply from $38.30$ (score of TC in Table~\ref{tab:study2_1}) to $23.0$. 

\begin{sloppypar}
Moreover, from the perspective of target environments, the results also show consistency with single-environment generalization. For target environments easy to generalization in single-environment scenarios, such as TextCraft and BabyAI, sequentially trained agents also tend to perform well. Specifically, agents pre-trained on WebShop (i.e., WS+SQ, WS+AW, WS+BA) achieve significant gains on TextCraft, increasing by $11.00$, $10.12$, and $15.62$ points, respectively. However, for environments challenging for generalization, like AlfWorld, sequential training yields limited benefits. For instance, all three agents utilizing TextCraft as the upstream environment (TC+WS, TC+SQ, TC+BA) suffer performance degradation on AlfWorld.
\end{sloppypar}

\begin{figure*}[t]
  \centering
  \begin{subfigure}{0.99\textwidth}
    \includegraphics[width=\linewidth]{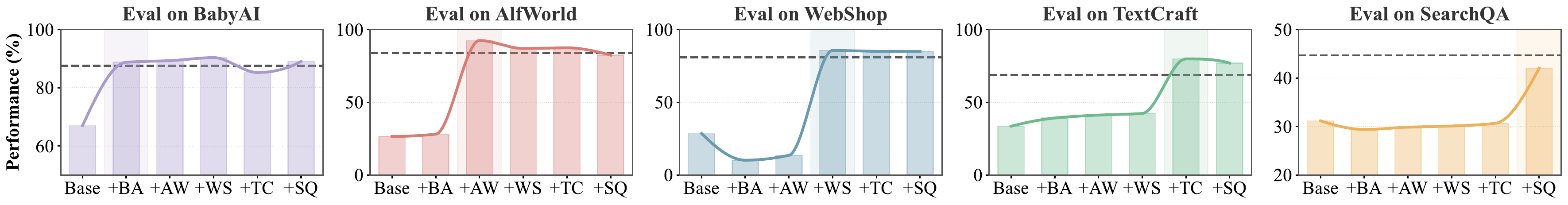}
    \caption{Sequential training of BabyAI, AlfWorld, WebShop, TextCraft, and SearchQA.}
  \end{subfigure}
\\
    \begin{subfigure}{0.99\textwidth}
    \includegraphics[width=\linewidth]{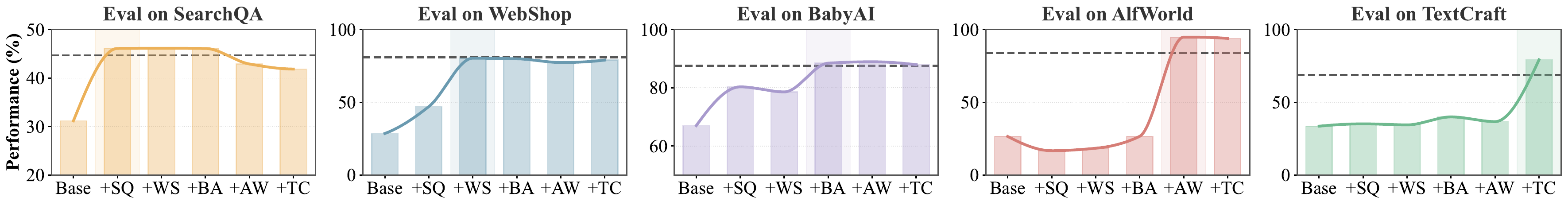}
    \caption{Sequential training of SearchQA, WebShop, BabyAI, AlfWorld, and TextCraft.}
  \end{subfigure}
  \caption{Training dynamics of sequential training across five environments. We present the results for representative sequence combinations, monitoring how performance on each environment changes as the agent is trained on different environments sequentially. The \emph{dashed lines} denote the performance achieved by \textbf{joint training on a mixture of data from all five environments}.}
  \label{fig:five_env}
\end{figure*}

\paragraph{Training order significantly affects generalization performance in held-out environments.}
As shown in Figure~\ref{fig:study3_generalization}, the training order exerts a substantial influence on generalization performance. For instance, when evaluated on TextCraft and AlfWorld, BA+SQ outperforms SQ+BA by $15.63$ and $10.00$ points, respectively. This represents a substantial margin, particularly for held-out scenarios. We attribute this phenomenon to the inherent task difficulty within each environment. As noted in Section \ref{sec:study2}, BabyAI provides detailed feedback, whereas SearchQA imposes strict constrains with limited feedback. Consequently, the BA+SQ order naturally creates an ``easy-to-hard'' curriculum, which in turn facilitates better generalization performance. 

\paragraph{Sequential training achieves performance comparable to joint training.}
Furthermore, we conduct sequential training over five environments in different orders. Figure~\ref{fig:five_env} illustrates the performance dynamics on each environment across the five training stages for two representative sequences; for comparison, joint-training results are indicated by dashed lines. The results show that sequential training achieves performance comparable to joint training, even after training on five distinct tasks. Overall, the final performance is insensitive to the training order, which we attribute to RFT’s ability to preserve previously acquired capabilities, consistent with findings in prior work.
For environments such as AlfWorld and SearchQA—to which other tasks struggle to generalize—relatively pronounced forgetting may occur over the course of long-term sequential training.

\section{Further Analysis and Discussion}\label{sec:discussion}

\paragraph{Failure Mode Analysis.} 

\begin{figure*}[t]
  \centering
  \includegraphics[width=\linewidth]{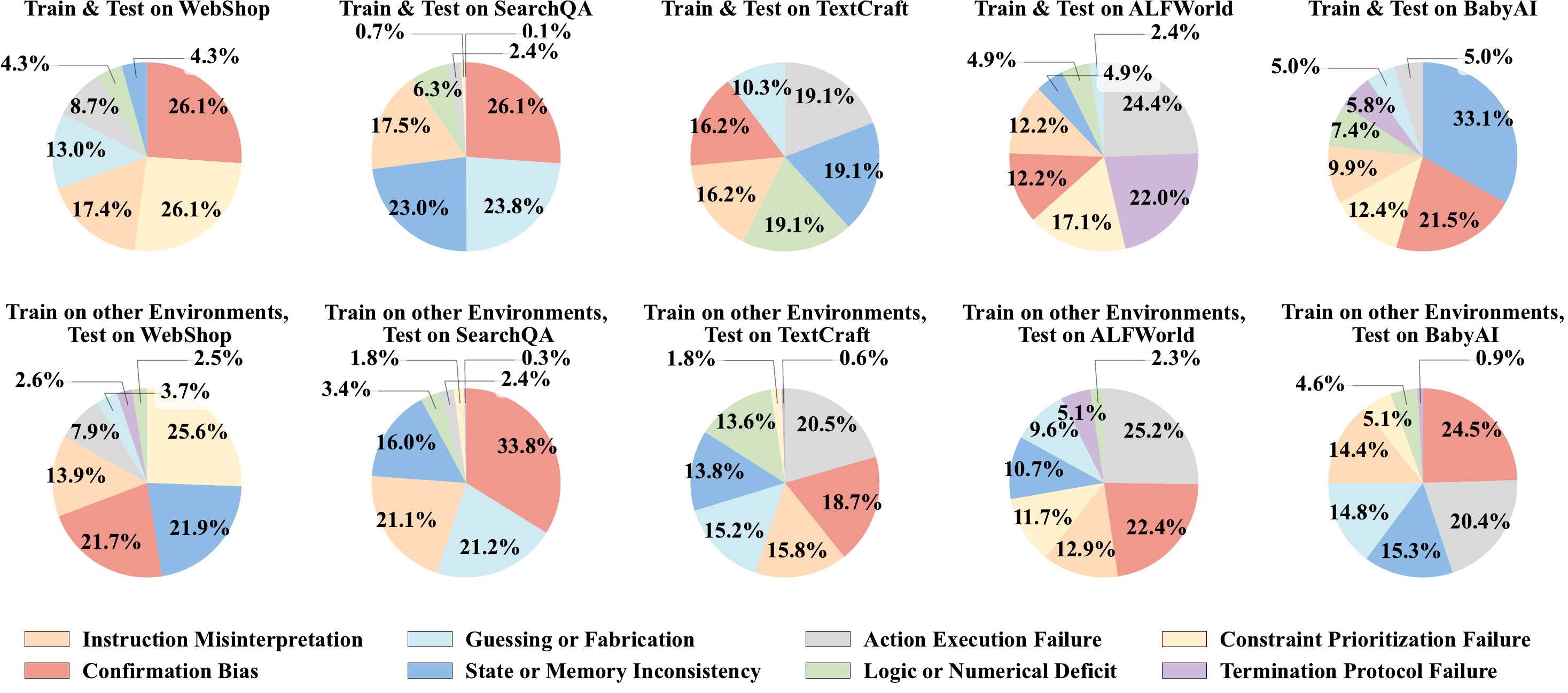}
  \caption{Comparison of failure mode distributions between in-environment and out-of-environment evaluation.}
  \label{fig:error_type}
\end{figure*}

We conduct a fine-grained error analysis on both intra/inter-environment evaluations, summarizing 8 common failure modes including \textit{instruction misinterpretation}, \textit{action execution failure}, and \textit{logical deficits}. Using GPT-5-mini, we systematically categorize the error trajectories across all scenarios. More detail about each failure pattern can be found in Appendix \ref{appendix:error_type}.

The results presented in Figure \ref{fig:error_type} reveal that while failure modes differ by environment, errors related to ``Confirmation Bias'' are prevalent ($>10\%$) across all scenarios. This suggests that after training, agents tend to exhibit overconfidence, neglecting further verification and lacking the capacity for self-reflection based on environmental feedback. Moreover, in SearchQA, errors categorized as ``Guessing or Fabrication'' are widespread among both held-in ($23.8\%$) and held-out ($21.2\%$) settings. This highlights a critical deficiency in tool utilization, which fundamentally constrains the agent's generalization potential. 

When the training and test environments differ, agents exhibit different failure modes. For example, in WebShop we observe an increase in the proportion of “State or Memory Inconsistency” errors, rising from $4.3\%$ in held-in to $21.9\%$ in held-out settings. This suggests that, when confronted with large volumes of information, the agent’s decision-making becomes less coherent, and it struggles to generalize its ability to extract salient information. A full environment-by-environment breakdown is provided in Figure~\ref{fig:error_matrix}.

\paragraph{Case Study.}

We perform a comprehensive analysis of environment generalization and present specific case studies in Appendix \ref{appendix:case_study}. Notably, Figure \ref{fig:case_webshop} illustrates the mechanism behind the successful transfer from SearchQA to WebShop. Specifically, the base agent tends to blindly input the entire instruction into the search bar, and struggles to accurately extract key information from the voluminous HTML content, resulting in the retrieval and selection of irrelevant items. In contrast, the agent trained on SearchQA learns to effectively search for key details and extract information, successfully selecting the correct item.

Furthermore, Figure~\ref{fig:case_searchqa} offers qualitative insight into why SearchQA generalization remains difficult, using an AlfWorld-trained agent as a case study. Although both agents fail initially, the SearchQA-trained agent iteratively refines its queries to become more specific, improving retrieval and ultimately recovering. By contrast, this query-refinement behavior does not reliably transfer: the AlfWorld-trained agent falls into a degenerate loop, repeatedly issuing near-duplicate searches, returning the same responses, and ultimately failing.

\section{Conclusion}
In this paper, we present a systematic study of how RFT affects the transfer and generalization of LLM agents for multi-turn decision-making. Through large-scale experiments along three complementary axes, we characterize when RFT generalizes within and across environments and identify generalization patterns. We further complement our quantitative results with failure mode analysis and qualitative case study to pinpoint where agents break down and what behaviors fail to transfer. Together, our findings offer practical guidance for training and evaluating agents under distribution shift, and we hope they inform the development of agents that generalize reliably in real-world deployments.

\bibliography{main}

@article{zhang2025reinforcement,
  title={Why Reinforcement Fine-Tuning Enables MLLMs Preserve Prior Knowledge Better: A Data Perspective},
  author={Zhang, Zhihao and Dong, Qiaole and Zhang, Qi and Zhao, Jun and Zhou, Enyu and Xi, Zhiheng and Jin, Senjie and Fan, Xiaoran and Zhou, Yuhao and Wu, Mingqi and others},
  journal={arXiv preprint arXiv:2506.23508},
  year={2025}
}

@inproceedings{DBLP:conf/nips/ChristianoLBMLA17,
  author       = {Paul F. Christiano and
                  Jan Leike and
                  Tom B. Brown and
                  Miljan Martic and
                  Shane Legg and
                  Dario Amodei},
  editor       = {Isabelle Guyon and
                  Ulrike von Luxburg and
                  Samy Bengio and
                  Hanna M. Wallach and
                  Rob Fergus and
                  S. V. N. Vishwanathan and
                  Roman Garnett},
  title        = {Deep Reinforcement Learning from Human Preferences},
  booktitle    = {Advances in Neural Information Processing Systems 30: Annual Conference
                  on Neural Information Processing Systems 2017, December 4-9, 2017,
                  Long Beach, CA, {USA}},
  pages        = {4299--4307},
  year         = {2017},
  bibsource    = {dblp computer science bibliography, https://dblp.org}
}

@inproceedings{DBLP:conf/nips/Ouyang0JAWMZASR22,
  author       = {Long Ouyang and
                  Jeffrey Wu and
                  Xu Jiang and
                  Diogo Almeida and
                  Carroll L. Wainwright and
                  Pamela Mishkin and
                  Chong Zhang and
                  Sandhini Agarwal and
                  Katarina Slama and
                  Alex Ray and
                  John Schulman and
                  Jacob Hilton and
                  Fraser Kelton and
                  Luke Miller and
                  Maddie Simens and
                  Amanda Askell and
                  Peter Welinder and
                  Paul F. Christiano and
                  Jan Leike and
                  Ryan Lowe},
  editor       = {Sanmi Koyejo and
                  S. Mohamed and
                  A. Agarwal and
                  Danielle Belgrave and
                  K. Cho and
                  A. Oh},
  title        = {Training language models to follow instructions with human feedback},
  booktitle    = {Advances in Neural Information Processing Systems 35: Annual Conference
                  on Neural Information Processing Systems 2022, NeurIPS 2022, New Orleans,
                  LA, USA, November 28 - December 9, 2022},
  year         = {2022},
  bibsource    = {dblp computer science bibliography, https://dblp.org}
}

@article{DBLP:journals/corr/abs-2212-08073,
  author       = {Yuntao Bai and
                  Saurav Kadavath and
                  Sandipan Kundu and
                  Amanda Askell and
                  Jackson Kernion and
                  Andy Jones and
                  Anna Chen and
                  Anna Goldie and
                  Azalia Mirhoseini and
                  Cameron McKinnon and
                  Carol Chen and
                  Catherine Olsson and
                  Christopher Olah and
                  Danny Hernandez and
                  Dawn Drain and
                  Deep Ganguli and
                  Dustin Li and
                  Eli Tran{-}Johnson and
                  Ethan Perez and
                  Jamie Kerr and
                  Jared Mueller and
                  Jeffrey Ladish and
                  Joshua Landau and
                  Kamal Ndousse and
                  Kamile Lukosiute and
                  Liane Lovitt and
                  Michael Sellitto and
                  Nelson Elhage and
                  Nicholas Schiefer and
                  Noem{\'{\i}} Mercado and
                  Nova DasSarma and
                  Robert Lasenby and
                  Robin Larson and
                  Sam Ringer and
                  Scott Johnston and
                  Shauna Kravec and
                  Sheer El Showk and
                  Stanislav Fort and
                  Tamera Lanham and
                  Timothy Telleen{-}Lawton and
                  Tom Conerly and
                  Tom Henighan and
                  Tristan Hume and
                  Samuel R. Bowman and
                  Zac Hatfield{-}Dodds and
                  Ben Mann and
                  Dario Amodei and
                  Nicholas Joseph and
                  Sam McCandlish and
                  Tom Brown and
                  Jared Kaplan},
  title        = {Constitutional {AI:} Harmlessness from {AI} Feedback},
  journal      = {CoRR},
  volume       = {abs/2212.08073},
  year         = {2022},
  url          = {https://doi.org/10.48550/arXiv.2212.08073},
  doi          = {10.48550/ARXIV.2212.08073},
  eprinttype    = {arXiv},
  eprint       = {2212.08073},
  bibsource    = {dblp computer science bibliography, https://dblp.org}
}

@article{DBLP:journals/corr/SchulmanWDRK17,
  author       = {John Schulman and
                  Filip Wolski and
                  Prafulla Dhariwal and
                  Alec Radford and
                  Oleg Klimov},
  title        = {Proximal Policy Optimization Algorithms},
  journal      = {CoRR},
  volume       = {abs/1707.06347},
  year         = {2017},
  url          = {http://arxiv.org/abs/1707.06347},
  eprinttype    = {arXiv},
  eprint       = {1707.06347},
  bibsource    = {dblp computer science bibliography, https://dblp.org}
}

@inproceedings{DBLP:conf/nips/RafailovSMMEF23,
  author       = {Rafael Rafailov and
                  Archit Sharma and
                  Eric Mitchell and
                  Christopher D. Manning and
                  Stefano Ermon and
                  Chelsea Finn},
  editor       = {Alice Oh and
                  Tristan Naumann and
                  Amir Globerson and
                  Kate Saenko and
                  Moritz Hardt and
                  Sergey Levine},
  title        = {Direct Preference Optimization: Your Language Model is Secretly a
                  Reward Model},
  booktitle    = {Advances in Neural Information Processing Systems 36: Annual Conference
                  on Neural Information Processing Systems 2023, NeurIPS 2023, New Orleans,
                  LA, USA, December 10 - 16, 2023},
  year         = {2023},
  bibsource    = {dblp computer science bibliography, https://dblp.org}
}

@article{DBLP:journals/corr/abs-2501-12948,
  author       = {DeepSeek{-}AI},
  title        = {DeepSeek-R1: Incentivizing Reasoning Capability in LLMs via Reinforcement
                  Learning},
  journal      = {CoRR},
  volume       = {abs/2501.12948},
  year         = {2025},
  url          = {https://doi.org/10.48550/arXiv.2501.12948},
  doi          = {10.48550/ARXIV.2501.12948},
  eprinttype    = {arXiv},
  eprint       = {2501.12948},
  bibsource    = {dblp computer science bibliography, https://dblp.org}
}

@article{DBLP:journals/corr/abs-2504-21801,
  author       = {Z. Z. Ren and
                  Zhihong Shao and
                  Junxiao Song and
                  Huajian Xin and
                  Haocheng Wang and
                  Wanjia Zhao and
                  Liyue Zhang and
                  Zhe Fu and
                  Qihao Zhu and
                  Dejian Yang and
                  Z. F. Wu and
                  Zhibin Gou and
                  Shirong Ma and
                  Hongxuan Tang and
                  Yuxuan Liu and
                  Wenjun Gao and
                  Daya Guo and
                  Chong Ruan},
  title        = {DeepSeek-Prover-V2: Advancing Formal Mathematical Reasoning via Reinforcement
                  Learning for Subgoal Decomposition},
  journal      = {CoRR},
  volume       = {abs/2504.21801},
  year         = {2025},
  url          = {https://doi.org/10.48550/arXiv.2504.21801},
  doi          = {10.48550/ARXIV.2504.21801},
  eprinttype    = {arXiv},
  eprint       = {2504.21801},
  bibsource    = {dblp computer science bibliography, https://dblp.org}
}

@article{guo2025ifdecorator,
  title={IFDECORATOR: Wrapping Instruction Following Reinforcement Learning with Verifiable Rewards},
  author={Guo, Xu and Liang, Tianyi and Jian, Tong and Yang, Xiaogui and Wu, Ling-I and Li, Chenhui and Lu, Zhihui and Guo, Qipeng and Chen, Kai},
  journal={arXiv preprint arXiv:2508.04632},
  year={2025}
}

@inproceedings{DBLP:conf/nips/MuHHAVKLBSW24,
  author       = {Tong Mu and
                  Alec Helyar and
                  Johannes Heidecke and
                  Joshua Achiam and
                  Andrea Vallone and
                  Ian Kivlichan and
                  Molly Lin and
                  Alex Beutel and
                  John Schulman and
                  Lilian Weng},
  editor       = {Amir Globersons and
                  Lester Mackey and
                  Danielle Belgrave and
                  Angela Fan and
                  Ulrich Paquet and
                  Jakub M. Tomczak and
                  Cheng Zhang},
  title        = {Rule Based Rewards for Language Model Safety},
  booktitle    = {Advances in Neural Information Processing Systems 38: Annual Conference
                  on Neural Information Processing Systems 2024, NeurIPS 2024, Vancouver,
                  BC, Canada, December 10 - 15, 2024},
  year         = {2024},
  bibsource    = {dblp computer science bibliography, https://dblp.org}
}

@inproceedings{zhai2025enhancing,
  author       = {Yuanzhao Zhai and
                  Tingkai Yang and
                  Kele Xu and
                  Dawei Feng and
                  Cheng Yang and
                  Bo Ding and
                  Huaimin Wang},
  editor       = {Toby Walsh and
                  Julie Shah and
                  Zico Kolter},
  title        = {Enhancing Decision-Making for {LLM} Agents via Step-Level Q-Value
                  Models},
  booktitle    = {AAAI-25, Sponsored by the Association for the Advancement of Artificial
                  Intelligence, February 25 - March 4, 2025, Philadelphia, PA, {USA}},
  pages        = {27161--27169},
  publisher    = {{AAAI} Press},
  year         = {2025},
  url          = {https://doi.org/10.1609/aaai.v39i25.34924},
  doi          = {10.1609/AAAI.V39I25.34924},
  bibsource    = {dblp computer science bibliography, https://dblp.org}
}

@article{DBLP:journals/corr/abs-2509-08755,
  author       = {Zhiheng Xi and
                  Jixuan Huang and
                  Chenyang Liao and
                  Baodai Huang and
                  Honglin Guo and
                  Jiaqi Liu and
                  Rui Zheng and
                  Junjie Ye and
                  Jiazheng Zhang and
                  Wenxiang Chen and
                  Wei He and
                  Yiwen Ding and
                  Guanyu Li and
                  Zehui Chen and
                  Zhengyin Du and
                  Xuesong Yao and
                  Yufei Xu and
                  Jiecao Chen and
                  Tao Gui and
                  Zuxuan Wu and
                  Qi Zhang and
                  Xuanjing Huang and
                  Yu{-}Gang Jiang},
  title        = {AgentGym-RL: Training {LLM} Agents for Long-Horizon Decision Making
                  through Multi-Turn Reinforcement Learning},
  journal      = {CoRR},
  volume       = {abs/2509.08755},
  year         = {2025},
  url          = {https://doi.org/10.48550/arXiv.2509.08755},
  doi          = {10.48550/ARXIV.2509.08755},
  eprinttype    = {arXiv},
  eprint       = {2509.08755},
  bibsource    = {dblp computer science bibliography, https://dblp.org}
}

@article{DBLP:journals/corr/abs-2502-01600,
  author       = {Kevin Chen and
                  Marco F. Cusumano{-}Towner and
                  Brody Huval and
                  Aleksei Petrenko and
                  Jackson Hamburger and
                  Vladlen Koltun and
                  Philipp Kr{\"{a}}henb{\"{u}}hl},
  title        = {Reinforcement Learning for Long-Horizon Interactive {LLM} Agents},
  journal      = {CoRR},
  volume       = {abs/2502.01600},
  year         = {2025},
  url          = {https://doi.org/10.48550/arXiv.2502.01600},
  doi          = {10.48550/ARXIV.2502.01600},
  eprinttype    = {arXiv},
  eprint       = {2502.01600},
  bibsource    = {dblp computer science bibliography, https://dblp.org}
}

@inproceedings{DBLP:conf/nips/FengHHLZZL24,
  author       = {Peiyuan Feng and
                  Yichen He and
                  Guanhua Huang and
                  Yuan Lin and
                  Hanchong Zhang and
                  Yuchen Zhang and
                  Hang Li},
  editor       = {Amir Globersons and
                  Lester Mackey and
                  Danielle Belgrave and
                  Angela Fan and
                  Ulrich Paquet and
                  Jakub M. Tomczak and
                  Cheng Zhang},
  title        = {{AGILE:} {A} Novel Reinforcement Learning Framework of {LLM} Agents},
  booktitle    = {Advances in Neural Information Processing Systems 38: Annual Conference
                  on Neural Information Processing Systems 2024, NeurIPS 2024, Vancouver,
                  BC, Canada, December 10 - 15, 2024},
  year         = {2024},
  bibsource    = {dblp computer science bibliography, https://dblp.org}
}

@article{DBLP:journals/corr/abs-2505-20732,
  author       = {Hanlin Wang and
                  Chak Tou Leong and
                  Jiashuo Wang and
                  Jian Wang and
                  Wenjie Li},
  title        = {{SPA-RL:} Reinforcing {LLM} Agents via Stepwise Progress Attribution},
  journal      = {CoRR},
  volume       = {abs/2505.20732},
  year         = {2025},
  url          = {https://doi.org/10.48550/arXiv.2505.20732},
  doi          = {10.48550/ARXIV.2505.20732},
  eprinttype    = {arXiv},
  eprint       = {2505.20732},
  bibsource    = {dblp computer science bibliography, https://dblp.org}
}

@inproceedings{wang2025offline,
  title={Offline reinforcement learning for llm multi-step reasoning},
  author={Wang, Huaijie and Hao, Shibo and Dong, Hanze and Zhang, Shenao and Bao, Yilin and Yang, Ziran and Wu, Yi},
  booktitle={Findings of the Association for Computational Linguistics: ACL 2025},
  pages={8881--8893},
  year={2025}
}

@inproceedings{DBLP:conf/iclr/ZhangKC25,
  author       = {Michael J. Q. Zhang and
                  W. Bradley Knox and
                  Eunsol Choi},
  title        = {Modeling Future Conversation Turns to Teach LLMs to Ask Clarifying
                  Questions},
  booktitle    = {The Thirteenth International Conference on Learning Representations,
                  {ICLR} 2025, Singapore, April 24-28, 2025},
  publisher    = {OpenReview.net},
  year         = {2025},
  url          = {https://openreview.net/forum?id=cwuSAR7EKd},
  bibsource    = {dblp computer science bibliography, https://dblp.org}
}

@article{ramrakhya2025grounding,
  title={Grounding Multimodal LLMs to Embodied Agents that Ask for Help with Reinforcement Learning},
  author={Ramrakhya, Ram and Chang, Matthew and Puig, Xavier and Desai, Ruta and Kira, Zsolt and Mottaghi, Roozbeh},
  journal={arXiv preprint arXiv:2504.00907},
  year={2025}
}

@article{DBLP:journals/corr/abs-2503-09516,
  author       = {Bowen Jin and
                  Hansi Zeng and
                  Zhenrui Yue and
                  Dong Wang and
                  Hamed Zamani and
                  Jiawei Han},
  title        = {Search-R1: Training LLMs to Reason and Leverage Search Engines with
                  Reinforcement Learning},
  journal      = {CoRR},
  volume       = {abs/2503.09516},
  year         = {2025},
  url          = {https://doi.org/10.48550/arXiv.2503.09516},
  doi          = {10.48550/ARXIV.2503.09516},
  eprinttype    = {arXiv},
  eprint       = {2503.09516},
  bibsource    = {dblp computer science bibliography, https://dblp.org}
}

@inproceedings{DBLP:conf/iclr/KumarZASCSBIBRZ25,
  author       = {Aviral Kumar and
                  Vincent Zhuang and
                  Rishabh Agarwal and
                  Yi Su and
                  John D. Co{-}Reyes and
                  Avi Singh and
                  Kate Baumli and
                  Shariq Iqbal and
                  Colton Bishop and
                  Rebecca Roelofs and
                  Lei M. Zhang and
                  Kay McKinney and
                  Disha Shrivastava and
                  Cosmin Paduraru and
                  George Tucker and
                  Doina Precup and
                  Feryal M. P. Behbahani and
                  Aleksandra Faust},
  title        = {Training Language Models to Self-Correct via Reinforcement Learning},
  booktitle    = {The Thirteenth International Conference on Learning Representations,
                  {ICLR} 2025, Singapore, April 24-28, 2025},
  publisher    = {OpenReview.net},
  year         = {2025},
  url          = {https://openreview.net/forum?id=CjwERcAU7w},
  bibsource    = {dblp computer science bibliography, https://dblp.org}
}

@article{DBLP:journals/corr/abs-2502-12853,
  author       = {Ruotian Ma and
                  Peisong Wang and
                  Cheng Liu and
                  Xingyan Liu and
                  Jiaqi Chen and
                  Bang Zhang and
                  Xin Zhou and
                  Nan Du and
                  Jia Li},
  title        = {S\({}^{\mbox{2}}\)R: Teaching LLMs to Self-verify and Self-correct
                  via Reinforcement Learning},
  journal      = {CoRR},
  volume       = {abs/2502.12853},
  year         = {2025},
  url          = {https://doi.org/10.48550/arXiv.2502.12853},
  doi          = {10.48550/ARXIV.2502.12853},
  eprinttype    = {arXiv},
  eprint       = {2502.12853},
  bibsource    = {dblp computer science bibliography, https://dblp.org}
}

@article{DBLP:journals/corr/abs-2511-01934,
  author       = {Yirong Zeng and
                  Xiao Ding and
                  Yutai Hou and
                  Yuxian Wang and
                  Li Du and
                  Juyi Dai and
                  Qiuyang Ding and
                  Duyu Tang and
                  Dandan Tu and
                  Weiwen Liu and
                  Bing Qin and
                  Ting Liu},
  title        = {Tool Zero: Training Tool-Augmented LLMs via Pure {RL} from Scratch},
  journal      = {CoRR},
  volume       = {abs/2511.01934},
  year         = {2025},
  url          = {https://doi.org/10.48550/arXiv.2511.01934},
  doi          = {10.48550/ARXIV.2511.01934},
  eprinttype    = {arXiv},
  eprint       = {2511.01934},
  bibsource    = {dblp computer science bibliography, https://dblp.org}
}

@article{DBLP:journals/corr/abs-2504-11536,
  author       = {Jiazhan Feng and
                  Shijue Huang and
                  Xingwei Qu and
                  Ge Zhang and
                  Yujia Qin and
                  Baoquan Zhong and
                  Chengquan Jiang and
                  Jinxin Chi and
                  Wanjun Zhong},
  title        = {ReTool: Reinforcement Learning for Strategic Tool Use in LLMs},
  journal      = {CoRR},
  volume       = {abs/2504.11536},
  year         = {2025},
  url          = {https://doi.org/10.48550/arXiv.2504.11536},
  doi          = {10.48550/ARXIV.2504.11536},
  eprinttype    = {arXiv},
  eprint       = {2504.11536},
  bibsource    = {dblp computer science bibliography, https://dblp.org}
}

@article{DBLP:journals/corr/abs-2510-21618,
  author       = {Xiaoxi Li and
                  Wenxiang Jiao and
                  Jiarui Jin and
                  Guanting Dong and
                  Jiajie Jin and
                  Yinuo Wang and
                  Hao Wang and
                  Yutao Zhu and
                  Ji{-}Rong Wen and
                  Yuan Lu and
                  Zhicheng Dou},
  title        = {DeepAgent: {A} General Reasoning Agent with Scalable Toolsets},
  journal      = {CoRR},
  volume       = {abs/2510.21618},
  year         = {2025},
  url          = {https://doi.org/10.48550/arXiv.2510.21618},
  doi          = {10.48550/ARXIV.2510.21618},
  eprinttype    = {arXiv},
  eprint       = {2510.21618},
  bibsource    = {dblp computer science bibliography, https://dblp.org}
}

@article{zeng2025evolving,
  title={Evolving LLMs' Self-Refinement Capability via Iterative Preference Optimization},
  author={Zeng, Yongcheng and Cui, Xinyu and Jin, Xuanfa and Liu, Guoqing and Sun, Zexu and Li, Dong and Yang, Ning and Hao, Jianye and Zhang, Haifeng and Wang, Jun},
  journal={arXiv preprint arXiv:2502.05605},
  year={2025}
}

@article{zhai2024fine,
  title={Fine-tuning large vision-language models as decision-making agents via reinforcement learning},
  author={Zhai, Simon and Bai, Hao and Lin, Zipeng and Pan, Jiayi and Tong, Peter and Zhou, Yifei and Suhr, Alane and Xie, Saining and LeCun, Yann and Ma, Yi and others},
  journal={Advances in neural information processing systems},
  volume={37},
  pages={110935--110971},
  year={2024}
}

@article{DBLP:journals/corr/abs-2403-02502,
  author       = {Yifan Song and
                  Da Yin and
                  Xiang Yue and
                  Jie Huang and
                  Sujian Li and
                  Bill Yuchen Lin},
  title        = {Trial and Error: Exploration-Based Trajectory Optimization for {LLM}
                  Agents},
  journal      = {CoRR},
  volume       = {abs/2403.02502},
  year         = {2024},
  url          = {https://doi.org/10.48550/arXiv.2403.02502},
  doi          = {10.48550/ARXIV.2403.02502},
  eprinttype    = {arXiv},
  eprint       = {2403.02502},
  bibsource    = {dblp computer science bibliography, https://dblp.org}
}

@article{tan2024cradle,
  title={Cradle: Empowering foundation agents towards general computer control},
  author={Tan, Weihao and Zhang, Wentao and Xu, Xinrun and Xia, Haochong and Ding, Ziluo and Li, Boyu and Zhou, Bohan and Yue, Junpeng and Jiang, Jiechuan and Li, Yewen and others},
  journal={arXiv preprint arXiv:2403.03186},
  year={2024}
}

@inproceedings{DBLP:conf/iclr/KumarRJ0L22,
  author       = {Ananya Kumar and
                  Aditi Raghunathan and
                  Robbie Matthew Jones and
                  Tengyu Ma and
                  Percy Liang},
  title        = {Fine-Tuning can Distort Pretrained Features and Underperform Out-of-Distribution},
  booktitle    = {The Tenth International Conference on Learning Representations, {ICLR}
                  2022, Virtual Event, April 25-29, 2022},
  publisher    = {OpenReview.net},
  year         = {2022},
  url          = {https://openreview.net/forum?id=UYneFzXSJWh},
  bibsource    = {dblp computer science bibliography, https://dblp.org}
}

@article{DBLP:journals/corr/abs-2308-08747,
  author       = {Yun Luo and
                  Zhen Yang and
                  Fandong Meng and
                  Yafu Li and
                  Jie Zhou and
                  Yue Zhang},
  title        = {An Empirical Study of Catastrophic Forgetting in Large Language Models
                  During Continual Fine-tuning},
  journal      = {CoRR},
  volume       = {abs/2308.08747},
  year         = {2023},
  url          = {https://doi.org/10.48550/arXiv.2308.08747},
  doi          = {10.48550/ARXIV.2308.08747},
  eprinttype    = {arXiv},
  eprint       = {2308.08747},
  bibsource    = {dblp computer science bibliography, https://dblp.org}
}

@inproceedings{DBLP:conf/iclr/KothaSR24,
  author       = {Suhas Kotha and
                  Jacob Mitchell Springer and
                  Aditi Raghunathan},
  title        = {Understanding Catastrophic Forgetting in Language Models via Implicit
                  Inference},
  booktitle    = {The Twelfth International Conference on Learning Representations,
                  {ICLR} 2024, Vienna, Austria, May 7-11, 2024},
  publisher    = {OpenReview.net},
  year         = {2024},
  url          = {https://openreview.net/forum?id=VrHiF2hsrm},
  bibsource    = {dblp computer science bibliography, https://dblp.org}
}

@article{DBLP:journals/corr/abs-2507-00432,
  author       = {Maggie Huan and
                  Yuetai Li and
                  Tuney Zheng and
                  Xiaoyu Xu and
                  Seungone Kim and
                  Minxin Du and
                  Radha Poovendran and
                  Graham Neubig and
                  Xiang Yue},
  title        = {Does Math Reasoning Improve General {LLM} Capabilities? Understanding
                  Transferability of {LLM} Reasoning},
  journal      = {CoRR},
  volume       = {abs/2507.00432},
  year         = {2025},
  url          = {https://doi.org/10.48550/arXiv.2507.00432},
  doi          = {10.48550/ARXIV.2507.00432},
  eprinttype    = {arXiv},
  eprint       = {2507.00432},
  bibsource    = {dblp computer science bibliography, https://dblp.org}
}

@article{DBLP:journals/corr/abs-2510-18874,
  author       = {Howard Chen and
                  Noam Razin and
                  Karthik Narasimhan and
                  Danqi Chen},
  title        = {Retaining by Doing: The Role of On-Policy Data in Mitigating Forgetting},
  journal      = {CoRR},
  volume       = {abs/2510.18874},
  year         = {2025},
  url          = {https://doi.org/10.48550/arXiv.2510.18874},
  doi          = {10.48550/ARXIV.2510.18874},
  eprinttype    = {arXiv},
  eprint       = {2510.18874},
  bibsource    = {dblp computer science bibliography, https://dblp.org}
}

@article{cheng2025revisiting,
  title={Revisiting Reinforcement Learning for LLM Reasoning from A Cross-Domain Perspective},
  author={Cheng, Zhoujun and Hao, Shibo and Liu, Tianyang and Zhou, Fan and Xie, Yutao and Yao, Feng and Bian, Yuexin and Zhuang, Yonghao and Dey, Nilabjo and Zha, Yuheng and others},
  journal={arXiv preprint arXiv:2506.14965},
  year={2025}
}

@article{DBLP:journals/corr/abs-2510-02230,
  author       = {Phuc Minh Nguyen and
                  Chinh D. La and
                  Duy M. H. Nguyen and
                  Nitesh V. Chawla and
                  Binh T. Nguyen and
                  Khoa D. Doan},
  title        = {The Reasoning Boundary Paradox: How Reinforcement Learning Constrains
                  Language Models},
  journal      = {CoRR},
  volume       = {abs/2510.02230},
  year         = {2025},
  url          = {https://doi.org/10.48550/arXiv.2510.02230},
  doi          = {10.48550/ARXIV.2510.02230},
  eprinttype    = {arXiv},
  eprint       = {2510.02230},
  bibsource    = {dblp computer science bibliography, https://dblp.org}
}

@article{DBLP:journals/corr/abs-2506-19733,
  author       = {Chuxuan Hu and
                  Yuxuan Zhu and
                  Antony Kellermann and
                  Caleb Biddulph and
                  Suppakit Waiwitlikhit and
                  Jason Benn and
                  Daniel Kang},
  title        = {Breaking Barriers: Do Reinforcement Post Training Gains Transfer To
                  Unseen Domains?},
  journal      = {CoRR},
  volume       = {abs/2506.19733},
  year         = {2025},
  url          = {https://doi.org/10.48550/arXiv.2506.19733},
  doi          = {10.48550/ARXIV.2506.19733},
  eprinttype    = {arXiv},
  eprint       = {2506.19733},
  bibsource    = {dblp computer science bibliography, https://dblp.org}
}

@article{sun2025omega,
  title={OMEGA: Can LLMs Reason Outside the Box in Math? Evaluating Exploratory, Compositional, and Transformative Generalization},
  author={Sun, Yiyou and Hu, Shawn and Zhou, Georgia and Zheng, Ken and Hajishirzi, Hannaneh and Dziri, Nouha and Song, Dawn},
  journal={arXiv preprint arXiv:2506.18880},
  year={2025}
}

@article{DBLP:journals/tist/ZhaoCYLDCWYD24,
  author       = {Haiyan Zhao and
                  Hanjie Chen and
                  Fan Yang and
                  Ninghao Liu and
                  Huiqi Deng and
                  Hengyi Cai and
                  Shuaiqiang Wang and
                  Dawei Yin and
                  Mengnan Du},
  title        = {Explainability for Large Language Models: {A} Survey},
  journal      = {{ACM} Trans. Intell. Syst. Technol.},
  volume       = {15},
  number       = {2},
  pages        = {20:1--20:38},
  year         = {2024},
  url          = {https://doi.org/10.1145/3639372},
  doi          = {10.1145/3639372},
  bibsource    = {dblp computer science bibliography, https://dblp.org}
}

@article{DBLP:journals/corr/abs-2510-23081,
  author       = {Chengying Tu and
                  Xuemiao Zhang and
                  Rongxiang Weng and
                  Rumei Li and
                  Chen Zhang and
                  Yang Bai and
                  Hongfei Yan and
                  Jingang Wang and
                  Xunliang Cai},
  title        = {A Survey on {LLM} Mid-training},
  journal      = {CoRR},
  volume       = {abs/2510.23081},
  year         = {2025},
  url          = {https://doi.org/10.48550/arXiv.2510.23081},
  doi          = {10.48550/ARXIV.2510.23081},
  eprinttype    = {arXiv},
  eprint       = {2510.23081},
  bibsource    = {dblp computer science bibliography, https://dblp.org}
}

@article{zhou2023webarena,
  title={Webarena: A realistic web environment for building autonomous agents},
  author={Zhou, Shuyan and Xu, Frank F and Zhu, Hao and Zhou, Xuhui and Lo, Robert and Sridhar, Abishek and Cheng, Xianyi and Ou, Tianyue and Bisk, Yonatan and Fried, Daniel and others},
  journal={arXiv preprint arXiv:2307.13854},
  year={2023}
}

@article{he2024webvoyager,
  title={Webvoyager: Building an end-to-end web agent with large multimodal models},
  author={He, Hongliang and Yao, Wenlin and Ma, Kaixin and Yu, Wenhao and Dai, Yong and Zhang, Hongming and Lan, Zhenzhong and Yu, Dong},
  journal={arXiv preprint arXiv:2401.13919},
  year={2024}
}

@article{jimenez2023swe,
  title={Swe-bench: Can language models resolve real-world github issues?},
  author={Jimenez, Carlos E and Yang, John and Wettig, Alexander and Yao, Shunyu and Pei, Kexin and Press, Ofir and Narasimhan, Karthik},
  journal={arXiv preprint arXiv:2310.06770},
  year={2023}
}

@inproceedings{yao2022react,
  title={React: Synergizing reasoning and acting in language models},
  author={Yao, Shunyu and Zhao, Jeffrey and Yu, Dian and Du, Nan and Shafran, Izhak and Narasimhan, Karthik R and Cao, Yuan},
  booktitle={The eleventh international conference on learning representations},
  year={2022}
}

@article{shinn2023reflexion,
  title={Reflexion: Language agents with verbal reinforcement learning},
  author={Shinn, Noah and Cassano, Federico and Gopinath, Ashwin and Narasimhan, Karthik and Yao, Shunyu},
  journal={Advances in Neural Information Processing Systems},
  volume={36},
  pages={8634--8652},
  year={2023}
}

@inproceedings{lightman2023let,
  title={Let's verify step by step},
  author={Lightman, Hunter and Kosaraju, Vineet and Burda, Yuri and Edwards, Harrison and Baker, Bowen and Lee, Teddy and Leike, Jan and Schulman, John and Sutskever, Ilya and Cobbe, Karl},
  booktitle={The Twelfth International Conference on Learning Representations},
  year={2023}
}

@article{liu2023agentbench,
  title={Agentbench: Evaluating llms as agents},
  author={Liu, Xiao and Yu, Hao and Zhang, Hanchen and Xu, Yifan and Lei, Xuanyu and Lai, Hanyu and Gu, Yu and Ding, Hangliang and Men, Kaiwen and Yang, Kejuan and others},
  journal={arXiv preprint arXiv:2308.03688},
  year={2023}
}

@article{xi2025rise,
  title={The rise and potential of large language model based agents: A survey},
  author={Xi, Zhiheng and Chen, Wenxiang and Guo, Xin and He, Wei and Ding, Yiwen and Hong, Boyang and Zhang, Ming and Wang, Junzhe and Jin, Senjie and Zhou, Enyu and others},
  journal={Science China Information Sciences},
  volume={68},
  number={2},
  pages={121101},
  year={2025},
  publisher={Springer}
}

@inproceedings{ruan2023tptu,
  title={Tptu: Task planning and tool usage of large language model-based ai agents},
  author={Ruan, Jingqing and Chen, Yihong and Zhang, Bin and Xu, Zhiwei and Bao, Tianpeng and Mao, Hangyu and Li, Ziyue and Zeng, Xingyu and Zhao, Rui and others},
  booktitle={NeurIPS 2023 Foundation Models for Decision Making Workshop},
  year={2023}
}

@article{li2025deepagent,
  title={Deepagent: A general reasoning agent with scalable toolsets},
  author={Li, Xiaoxi and Jiao, Wenxiang and Jin, Jiarui and Dong, Guanting and Jin, Jiajie and Wang, Yinuo and Wang, Hao and Zhu, Yutao and Wen, Ji-Rong and Lu, Yuan and others},
  journal={arXiv preprint arXiv:2510.21618},
  year={2025}
}

@article{sutton1999policy,
  title={Policy gradient methods for reinforcement learning with function approximation},
  author={Sutton, Richard S and McAllester, David and Singh, Satinder and Mansour, Yishay},
  journal={Advances in neural information processing systems},
  volume={12},
  year={1999}
}

@article{ouyang2022training,
  title={Training language models to follow instructions with human feedback},
  author={Ouyang, Long and Wu, Jeffrey and Jiang, Xu and Almeida, Diogo and Wainwright, Carroll and Mishkin, Pamela and Zhang, Chong and Agarwal, Sandhini and Slama, Katarina and Ray, Alex and others},
  journal={Advances in neural information processing systems},
  volume={35},
  pages={27730--27744},
  year={2022}
}

@article{bai2022constitutional,
  title={Constitutional ai: Harmlessness from ai feedback},
  author={Bai, Yuntao and Kadavath, Saurav and Kundu, Sandipan and Askell, Amanda and Kernion, Jackson and Jones, Andy and Chen, Anna and Goldie, Anna and Mirhoseini, Azalia and McKinnon, Cameron and others},
  journal={arXiv preprint arXiv:2212.08073},
  year={2022}
}

@article{uesato2022solving,
  title={Solving math word problems with process-and outcome-based feedback},
  author={Uesato, Jonathan and Kushman, Nate and Kumar, Ramana and Song, Francis and Siegel, Noah and Wang, Lisa and Creswell, Antonia and Irving, Geoffrey and Higgins, Irina},
  journal={arXiv preprint arXiv:2211.14275},
  year={2022}
}

@article{yao2022webshop,
  title={Webshop: Towards scalable real-world web interaction with grounded language agents},
  author={Yao, Shunyu and Chen, Howard and Yang, John and Narasimhan, Karthik},
  journal={Advances in Neural Information Processing Systems},
  volume={35},
  pages={20744--20757},
  year={2022}
}

@article{dunn2017searchqa,
  title={Searchqa: A new q\&a dataset augmented with context from a search engine},
  author={Dunn, Matthew and Sagun, Levent and Higgins, Mike and Guney, V Ugur and Cirik, Volkan and Cho, Kyunghyun},
  journal={arXiv preprint arXiv:1704.05179},
  year={2017}
}

@article{shridhar2020alfworld,
  title={Alfworld: Aligning text and embodied environments for interactive learning},
  author={Shridhar, Mohit and Yuan, Xingdi and C{\^o}t{\'e}, Marc-Alexandre and Bisk, Yonatan and Trischler, Adam and Hausknecht, Matthew},
  journal={arXiv preprint arXiv:2010.03768},
  year={2020}
}

@article{chevalier2018babyai,
  title={Babyai: A platform to study the sample efficiency of grounded language learning},
  author={Chevalier-Boisvert, Maxime and Bahdanau, Dzmitry and Lahlou, Salem and Willems, Lucas and Saharia, Chitwan and Nguyen, Thien Huu and Bengio, Yoshua},
  journal={arXiv preprint arXiv:1810.08272},
  year={2018}
}

@article{sanghi2022textcraft,
  author       = {Aditya Sanghi and
                  Rao Fu and
                  Vivian Liu and
                  Karl D. D. Willis and
                  Hooman Shayani and
                  Amir Hosein Khasahmadi and
                  Srinath Sridhar and
                  Daniel Ritchie},
  title        = {TextCraft: Zero-Shot Generation of High-Fidelity and Diverse Shapes
                  from Text},
  journal      = {CoRR},
  volume       = {abs/2211.01427},
  year         = {2022},
  url          = {https://doi.org/10.48550/arXiv.2211.01427},
  doi          = {10.48550/ARXIV.2211.01427},
  eprinttype    = {arXiv},
  eprint       = {2211.01427},
  bibsource    = {dblp computer science bibliography, https://dblp.org}
}

@inproceedings{xi2025agentgym,
  title={Agentgym: Evaluating and training large language model-based agents across diverse environments},
  author={Xi, Zhiheng and Ding, Yiwen and Chen, Wenxiang and Hong, Boyang and Guo, Honglin and Wang, Junzhe and Guo, Xin and Yang, Dingwen and Liao, Chenyang and He, Wei and others},
  booktitle={Proceedings of the 63rd Annual Meeting of the Association for Computational Linguistics (Volume 1: Long Papers)},
  pages={27914--27961},
  year={2025}
}

@misc{qwen2.5,
    title = {Qwen2.5: A Party of Foundation Models},
    url = {https://qwenlm.github.io/blog/qwen2.5/},
    author = {Qwen Team},
    month = {September},
    year = {2024}
}

@article{li2023tool,
  title={Tool-augmented reward modeling},
  author={Li, Lei and Chai, Yekun and Wang, Shuohuan and Sun, Yu and Tian, Hao and Zhang, Ningyu and Wu, Hua},
  journal={arXiv preprint arXiv:2310.01045},
  year={2023}
}

@article{mai2025agent,
  title={Agent rl scaling law: Agent rl with spontaneous code execution for mathematical problem solving},
  author={Mai, Xinji and Xu, Haotian and Li, Zhong-Zhi and Wang, Weinong and Hu, Jian and Zhang, Yingying and Zhang, Wenqiang and others},
  journal={arXiv preprint arXiv:2505.07773},
  year={2025}
}

@article{deng2023mind2web,
  title={Mind2web: Towards a generalist agent for the web},
  author={Deng, Xiang and Gu, Yu and Zheng, Boyuan and Chen, Shijie and Stevens, Sam and Wang, Boshi and Sun, Huan and Su, Yu},
  journal={Advances in Neural Information Processing Systems},
  volume={36},
  pages={28091--28114},
  year={2023}
}

@article{zan2025multi,
  title={Multi-swe-bench: A multilingual benchmark for issue resolving},
  author={Zan, Daoguang and Huang, Zhirong and Liu, Wei and Chen, Hanwu and Zhang, Linhao and Xin, Shulin and Chen, Lu and Liu, Qi and Zhong, Xiaojian and Li, Aoyan and others},
  journal={arXiv preprint arXiv:2504.02605},
  year={2025}
}

@article{merrill2026terminal,
  title={Terminal-Bench: Benchmarking Agents on Hard, Realistic Tasks in Command Line Interfaces},
  author={Merrill, Mike A and Shaw, Alexander G and Carlini, Nicholas and Li, Boxuan and Raj, Harsh and Bercovich, Ivan and Shi, Lin and Shin, Jeong Yeon and Walshe, Thomas and Buchanan, E Kelly and others},
  journal={arXiv preprint arXiv:2601.11868},
  year={2026}
}

@inproceedings{zhang2025learning,
  title={Learning like humans: Advancing llm reasoning capabilities via adaptive difficulty curriculum learning and expert-guided self-reformulation},
  author={Zhang, Enci and Yan, Xingang and Lin, Wei and Zhang, Tianxiang and Qianchun, Lu},
  booktitle={Proceedings of the 2025 Conference on Empirical Methods in Natural Language Processing},
  pages={6630--6644},
  year={2025}
}

@article{qi2024webrl,
  title={Webrl: Training llm web agents via self-evolving online curriculum reinforcement learning},
  author={Qi, Zehan and Liu, Xiao and Iong, Iat Long and Lai, Hanyu and Sun, Xueqiao and Zhao, Wenyi and Yang, Yu and Yang, Xinyue and Sun, Jiadai and Yao, Shuntian and others},
  journal={arXiv preprint arXiv:2411.02337},
  year={2024}
}

@article{nakano2021webgpt,
  title={Webgpt: Browser-assisted question-answering with human feedback},
  author={Nakano, Reiichiro and Hilton, Jacob and Balaji, Suchir and Wu, Jeff and Ouyang, Long and Kim, Christina and Hesse, Christopher and Jain, Shantanu and Kosaraju, Vineet and Saunders, William and others},
  journal={arXiv preprint arXiv:2112.09332},
  year={2021}
}

@article{song2024trial,
  title={Trial and error: Exploration-based trajectory optimization for llm agents},
  author={Song, Yifan and Yin, Da and Yue, Xiang and Huang, Jie and Li, Sujian and Lin, Bill Yuchen},
  journal={arXiv preprint arXiv:2403.02502},
  year={2024}
}

@inproceedings{bengio2009curriculum,
  title={Curriculum learning},
  author={Bengio, Yoshua and Louradour, J{\'e}r{\^o}me and Collobert, Ronan and Weston, Jason},
  booktitle={Proceedings of the 26th annual international conference on machine learning},
  pages={41--48},
  year={2009}
}

@article{mukherjee2023orca,
  title={Orca: Progressive learning from complex explanation traces of gpt-4},
  author={Mukherjee, Subhabrata and Mitra, Arindam and Jawahar, Ganesh and Agarwal, Sahaj and Palangi, Hamid and Awadallah, Ahmed},
  journal={arXiv preprint arXiv:2306.02707},
  year={2023}
}

@article{cui2025entropy,
  title={The entropy mechanism of reinforcement learning for reasoning language models},
  author={Cui, Ganqu and Zhang, Yuchen and Chen, Jiacheng and Yuan, Lifan and Wang, Zhi and Zuo, Yuxin and Li, Haozhan and Fan, Yuchen and Chen, Huayu and Chen, Weize and others},
  journal={arXiv preprint arXiv:2505.22617},
  year={2025}
}

@book{sutton1998reinforcement,
  author       = {Richard S. Sutton and
                  Andrew G. Barto},
  title        = {Reinforcement learning - an introduction},
  series       = {Adaptive computation and machine learning},
  publisher    = {{MIT} Press},
  year         = {1998},
  url          = {http://www.incompleteideas.net/book/first/the-book.html},
  isbn         = {978-0-262-19398-6},
  bibsource    = {dblp computer science bibliography, https://dblp.org}
}

@article{yu2025dapo,
  title={Dapo: An open-source llm reinforcement learning system at scale},
  author={Yu, Qiying and Zhang, Zheng and Zhu, Ruofei and Yuan, Yufeng and Zuo, Xiaochen and Yue, Yu and Dai, Weinan and Fan, Tiantian and Liu, Gaohong and Liu, Lingjun and others},
  journal={arXiv preprint arXiv:2503.14476},
  year={2025}
}

@article{hu2025reinforce++,
  title={Reinforce++: A simple and efficient approach for aligning large language models},
  author={Hu, Jian},
  journal={arXiv preprint arXiv:2501.03262},
  year={2025}
}

@article{wang2023voyager,
  title={Voyager: An open-ended embodied agent with large language models},
  author={Wang, Guanzhi and Xie, Yuqi and Jiang, Yunfan and Mandlekar, Ajay and Xiao, Chaowei and Zhu, Yuke and Fan, Linxi and Anandkumar, Anima},
  journal={arXiv preprint arXiv:2305.16291},
  year={2023}
}

@article{xu2023wizardlm,
  title={Wizardlm: Empowering large language models to follow complex instructions},
  author={Xu, Can and Sun, Qingfeng and Zheng, Kai and Geng, Xiubo and Zhao, Pu and Feng, Jiazhan and Tao, Chongyang and Jiang, Daxin},
  journal={arXiv preprint arXiv:2304.12244},
  year={2023}
}

@article{tian2024reinforcement,
  title={Reinforcement learning with adaptive regularization for safe control of critical systems},
  author={Tian, Haozhe and Hamedmoghadam, Homayoun and Shorten, Robert and Ferraro, Pietro},
  journal={Advances in Neural Information Processing Systems},
  volume={37},
  pages={2528--2557},
  year={2024}
}

@article{wang2024generalization,
  title={Generalization vs Memorization: Tracing Language Models' Capabilities Back to Pretraining Data},
  author={Wang, Xinyi and Antoniades, Antonis and Elazar, Yanai and Amayuelas, Alfonso and Albalak, Alon and Zhang, Kexun and Wang, William Yang},
  journal={arXiv preprint arXiv:2407.14985},
  year={2024}
}

@article{chu2025sft,
  title={Sft memorizes, rl generalizes: A comparative study of foundation model post-training},
  author={Chu, Tianzhe and Zhai, Yuexiang and Yang, Jihan and Tong, Shengbang and Xie, Saining and Schuurmans, Dale and Le, Quoc V and Levine, Sergey and Ma, Yi},
  journal={arXiv preprint arXiv:2501.17161},
  year={2025}
}

\clearpage
\newpage

\appendix
\section*{\centering \LARGE{Appendix}}

\section{Limitations and Future Work}
This paper adopts a new perspective to investigate how reinforcement fine-tuning affects the generalization ability of LLM agents.
Our experiments focus on the Qwen2.5 family and yield extensive empirical results together with substantive insights. Despite this progress, important challenges remain. First, we rely on default training/evaluation protocols and hyperparameter choices, without exhaustive tuning; more careful optimization may bring new findings. 
Second, we adopt GRPO—the most commonly used RL algorithm for LLM agents at present—rather than other approaches such as REINFORCE++ \citep{hu2025reinforce++} or DAPO \citep{yu2025dapo}; we leave these for future work.
Additionally, due to computational constraints, our sequential multi-environment training study does not enumerate all possible environment orderings; instead, we evaluate a small set of representative sequence. Although these experiments already reveal important insights, a broader and more fine-grained investigation of ordering effects is left for future work and may further substantiate our findings.

\section{Dataset Details}\label{appendix:environments}

We select five representative agent environments, including WebShop \cite{yao2022webshop}, SearchQA \cite{dunn2017searchqa}, TextCraft \cite{sanghi2022textcraft}, AlfWorld \cite{shridhar2020alfworld}, and BabyAI \cite{chevalier2018babyai}. The characteristics of each environment are shown in Table \ref{tab:compare_environments} in Section \ref{sec:experiment_setting}. Here, we provide their detailed types and action spaces in Table \ref{tab:action_space}.

\begin{table}[h]
  \caption{Detailed action spaces for each environment.}
  \vspace{-15pt}
  \label{tab:action_space}
  \begin{center}
    \begin{small}
        \begin{tabularx}{\textwidth}{ccX}
          \toprule
          \textbf{Environment}  & \textbf{Types} & \textbf{Action Spaces} \\
          \midrule
          WebShop & \emph{Web Navigation} & \emph{search, click} \\
          SearchQA & \emph{Q\&A Search} & \emph{search, answer} \\
          TextCraft & \emph{Text-based Game} & \emph{get, craft, inventory } \\
          AlfWorld & \emph{Household} & \emph{go to, open, close, take from, put in/on, use, heat, cool, clean, slice, inventory, look, examine} \\
          BabyAI & \emph{Embodied} & \emph{turn right, turn left, move forward, go to, go through, toggle and go through, toggle, pickup, drop, check available actions} \\
          \bottomrule
        \end{tabularx}
        \vspace{-6pt}
    \end{small}
  \end{center}
\end{table}

Our data is sourced from AgentGym \cite{xi2025agentgym}. Following practice in previous work \citep{bengio2009curriculum, mukherjee2023orca}, we categorize the tasks $\mathcal{U}$ into \emph{easy} and \emph{hard} difficulty levels (denoted as $\mathcal{U}_\text{easy}$ and $\mathcal{U}_\text{hard}$) based on the \texttt{avg@8} results of the Qwen2.5-7B-Instruct model, while ensuring a balanced distribution of data between the two difficulty levels. The same categorization is applied to the test set.
Detailed statistics for each environment are provided in Table~\ref{tab:environments}.

\begin{table}[h]
  \caption{Detailed data statistics for each environment.}
  \vspace{-15pt}
  \label{tab:environments}
  \begin{center}
    \begin{small}
        \begin{tabular}{crrrrrr}
          \toprule
          \multirow{2}{*}{\textbf{Environment}}  & \multicolumn{3}{c}{\textbf{Training dataset}} & \multicolumn{3}{c}{\textbf{Testing dataset}}  \\
          & \emph{easy} & \emph{hard} & \emph{all} & \emph{easy} & \emph{hard} & \emph{all} \\
          \midrule
          WebShop & $2104$ & $1826$ & $3930$ & $84$ & $116$ & $200$ \\
          SearchQA & $1960$ & $2040$ & $4000$ & $120$ & $280$ & $400$ \\
          TextCraft & $235$ & $209$ & $444$ & $57$ & $43$ & $100$ \\
          AlfWorld & $1267$ & $1153$ & $2420$ & $55$ & $145$ & $200$ \\
          BabyAI & $398$ & $412$ & $810$ & $52$ & $38$ & $90$ \\
          \bottomrule
        \end{tabular}
        \vspace{-6pt}
    \end{small}
  \end{center}
\end{table}

\section{Detailed algorithm of GRPO.}\label{appendix:grpo}

Group Relative Policy Optimization (GRPO) is an efficient online reinforcement learning algorithm tailored for LLMs. It  eliminates the need for a separate critic network typically required in PPO, thereby reducing computational overhead. Instead, GRPO estimates the baseline using \textit{group relative advantages} to significantly reduce gradient variance.

\begin{sloppypar}
Specifically, for each input query $q$ (derived from $u$), GRPO samples a group of outputs $\{y_1, y_2, \dots, y_G\}$ from the old policy $\pi_{\theta_{old}}$ and obtains their corresponding rewards $\{R_1, R_2, \dots, R_G\}$. The advantage $A_i$ for each output is calculated by normalizing the rewards within the group:
\begin{equation}
    A_i = \frac{r_i - \text{mean}(\{R_1, \dots, R_G\})}{\text{std}(\{R_1, \dots, R_G\})}
\end{equation}
Finally, GRPO updates the policy by maximizing the following surrogate objective, which incorporates PPO-style clipping \citep{DBLP:journals/corr/SchulmanWDRK17} and a KL divergence penalty to ensure training stability:
\end{sloppypar}
\begin{equation}
\begin{aligned}
\mathcal{J}_{\text{GRPO}}(\theta) = {} 
\mathbb{E}\Bigg[
\frac{1}{G}\sum_{i=1}^G \Big(
\min\!\big(
r_i(\theta) A_i,\;\text{clip}\big(r_i(\theta), 1-\epsilon, 1+\epsilon\big) A_i
\big)
- \beta \mathbb{D}_{\mathrm{KL}}
\Big)
\Bigg],
\end{aligned}
\end{equation}
where $r_i(\theta) = \frac{\pi_{\theta}(y_i|q)}{\pi_{\theta_{old}}(y_i|q)}$ denotes the probability ratio between the new and old policies, $\{y_i\}_{i=1}^G \sim \pi_{\theta_{\text{old}}}(q)$ represents the outputs $y_i$ sampled from the old policy $\pi_{\theta_{\text{old}}}$ given the query $q$, $\epsilon$ is the clipping parameter, and $\beta$ is the coefficient for the KL divergence term.

\section{Detailed Results of Sequential Cross-Environment Training}\label{appendix:study3}

\begin{figure}[h]
  \centering
  \includegraphics[width=\textwidth]{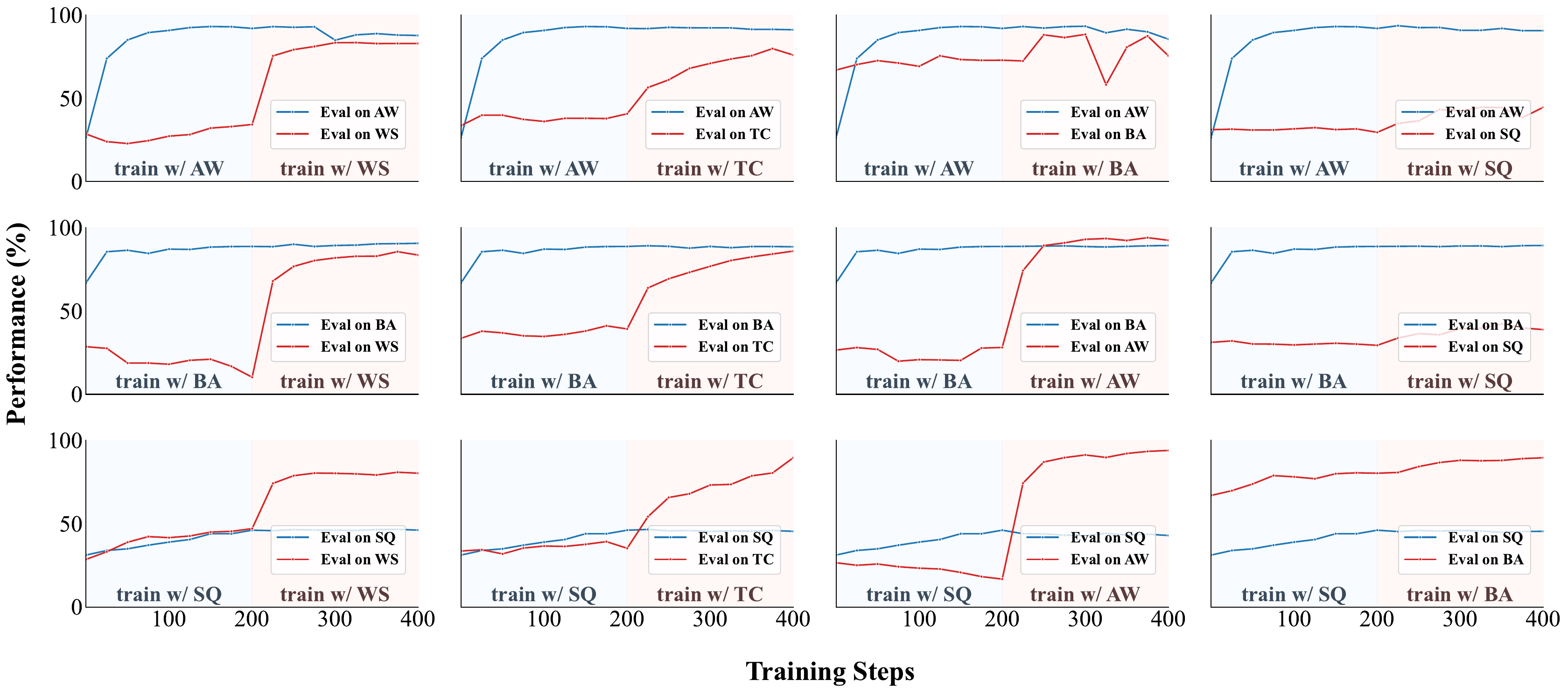}

  \caption{Training dynamics of forgetting and transfer in other sequential two-stage cross-environment training with Qwen2.5-7B-Instruct, where \textcolor{mymyblue}{blue} and \textcolor{mymyred}{red} denote the upstream environment and the downstream environment, respectively.}
  \label{fig:study3_dynamics_others}
\end{figure}

Figure \ref{fig:study3_dynamics} in Section \ref{sec:study3} illustrates the training dynamics for 8 two-stage sequential training configurations. Here, we present the dynamics for the remaining 12 pairs in Figure \ref{fig:study3_dynamics_others}. Additionally, detailed final results are reported in Table \ref{tab:study3}.

\begin{table*}[ht]
  \caption{Results of sequential training across different environments with Qwen2.5-7B-Instruct model.}
  \vspace{-15pt}
  \label{tab:study3}
  \begin{center}
    \begin{small}
      \begin{sc}
        \begin{tabular}{llccccc}
          \toprule
          \textbf{Upstream} & \textbf{Downstream} & \textbf{WebShop} & \textbf{SearchQA} & \textbf{TextCraft} & \textbf{AlfWorld} & \textbf{BabyAI} \\
          \midrule
          \rowcolor{gray!10}\multicolumn{2}{c}{base model} & $28.59$ & $31.19$ & $33.63$ & $26.56$ & $67.00$ \\
          \midrule
          \rowcolor{blue!5}WebShop & & $86.50$ & $33.28$ & $40.75$ & $24.13$ & $79.21$ \\
          & + SearchQA & $85.08$ & $41.44$ & $44.63$ & $17.81$ & $78.88$ \\
          & + TextCraft & $86.32$ & $36.81$ & $82.50$ & $26.63$ & $81.68$ \\
          & + AlfWorld & $85.99$ & $34.22$ & $43.75$ & $90.69$ & $76.34$ \\
          & + BabyAI & $86.38$ & $33.97$ & $49.25$ & $25.50$ & $87.74$ \\
          \midrule
          \rowcolor{blue!5}SearchQA & & $47.07$ & $46.12$ & $35.25$ & $16.75$ & $80.33$ \\
          & + WebShop & $80.35$ & $46.16$ & $34.50$ & $18.44$ & $78.59$ \\
          & + TextCraft & $49.31$ & $46.19$ & $78.63$ & $26.06$ & $78.03$ \\
          & + AlfWorld & $39.42$ & $42.92$ & $41.00$ & $94.00$ & $81.29$ \\
          & + BabyAI & $44.36$ & $45.44$ & $25.00$ & $14.06$ & $89.62$ \\
          \midrule
          \rowcolor{blue!5}TextCraft & & $38.30$ & $32.19$ & $80.88$ & $31.50$ & $77.95$ \\
          & + WebShop & $84.33$ & $34.59$ & $77.00$ & $18.31$ & $78.44$ \\
          & + SearchQA & $39.17$ & $43.16$ & $71.63$ & $19.69$ & $78.97$ \\
          & + AlfWorld & $42.04$ & $31.56$ & $76.50$ & $90.38$ & $62.19$ \\
          & + BabyAI & $23.00$ & $33.37$ & $74.88$ & $22.81$ & $87.55$ \\
          \midrule
          \rowcolor{blue!5}AlfWorld & & $34.31$ & $29.59$ & $36.13$ & $92.00$ & $72.91$ \\
          & + WebShop & $82.95$ & $31.50$ & $35.88$ & $87.75$ & $49.65$ \\
          & + SearchQA & $53.04$ & $44.66$ & $43.50$ & $90.63$ & $77.86$ \\
          & + TextCraft & $50.31$ & $30.81$ & $76.00$ & $91.19$ & $77.01$ \\
          & + BabyAI & $25.68$ & $28.31$ & $39.25$ & $85.56$ & $75.59$ \\
          \midrule
          \rowcolor{blue!5}BabyAI & & $10.25$ & $29.41$ & $39.25$ & $28.13$ & $88.79$ \\
          & + WebShop & $83.60$ & $30.19$ & $36.38$ & $33.38$ & $90.66$ \\
          & + SearchQA & $33.71$ & $38.84$ & $40.63$ & $24.06$ & $89.37$ \\
          & + TextCraft & $17.95$ & $33.37$ & $86.00$ & $29.31$ & $88.60$ \\
          & + AlfWorld & $13.60$ & $29.88$ & $41.25$ & $92.50$ & $89.32$ \\
          \bottomrule
        \end{tabular}
      \end{sc}
    \end{small}
  \end{center}
\end{table*}

\section{Detailed Results of average turns and generated tokens in different environment}\label{appendix:avg_turn_and_token}

\begin{figure*}[t]
  \centering
  \includegraphics[width=\textwidth]{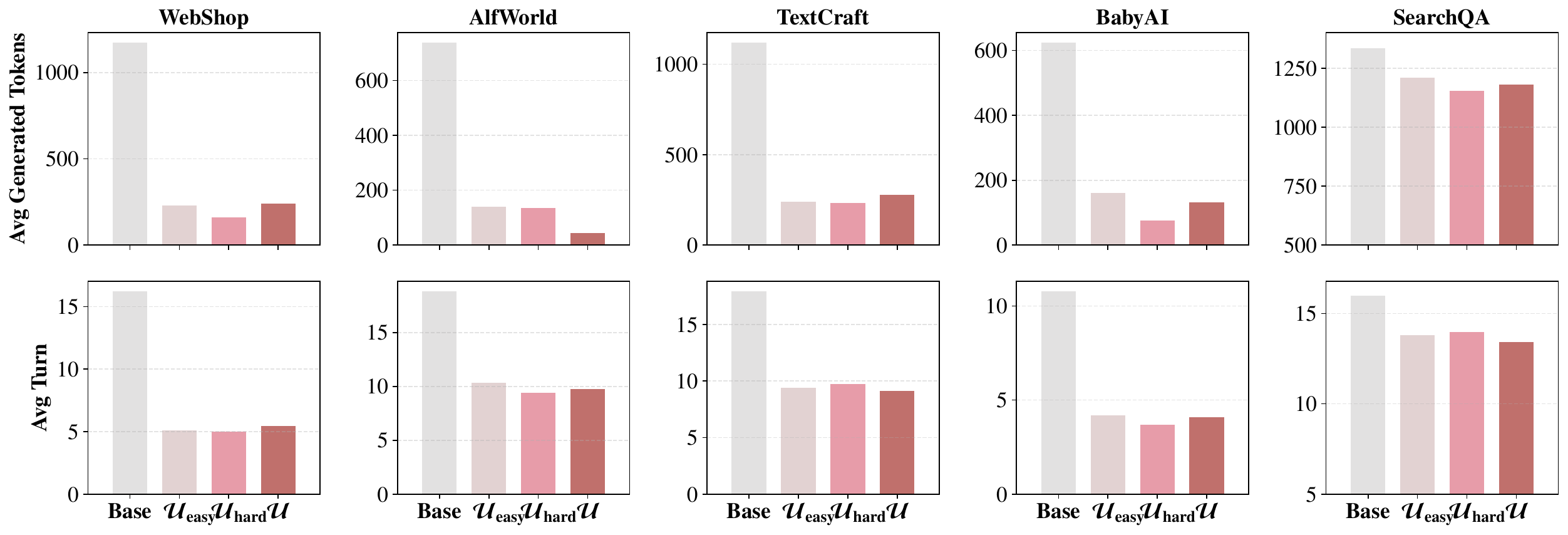}
  \caption{Average generated tokens and average turn across different environments for Qwen2.5-3B-Instruct model trained with varying difficulties.}
  \label{fig:avg_round_different_env}
\end{figure*}

Figure~\ref{fig:study1} in section~\ref{sec:study1} reports the average number of interaction turns and generated tokens across different environments. The more detailed results are reported in Figure~\ref{fig:avg_round_different_env}.

\section{Failure Mode Analysis}\label{appendix:error_type}

Through detailed analysis, we summarize 8 common error types:

\begin{enumerate}
    \item \textbf{Instruction Misinterpretation}: The agent fails to understand the prompt correctly, does not follow the instructions as indicated, or generates the wrong answer format.
    
    \item \textbf{Guessing or Fabrication}: The agent relies too heavily on internal parameters or makes unsupported guesses, instead of utilizing environmental tools to gather necessary external information.

    \item \textbf{Action Execution Failure}: The agent generates actions that cannot be interpreted by the environment or do not correspond to specific objects within the environment.
    
    \item \textbf{Constraint Prioritization Failure}: The agent recognizes multiple constraints in the instructions but fails to evaluate their relative importance correctly, resulting in the violation of core constraints in favor of secondary ones.

    \item \textbf{Confirmation Bias}: The agent becomes confident that it has found the correct answer or completed key steps, but does not proceed with further verification. Alternatively, it may forcefully apply other clues or validate erroneous information from the prompt.
    
    \item \textbf{State or Memory Inconsistency}: The agent exhibits contradictions over time, forgetting tools it has recently invoked or results it has obtained. Completed steps may be repeated unnecessarily.

    \item \textbf{Logic or Numerical Deficit}: The agent makes errors in logical or numerical reasoning.
    
    \item \textbf{Termination Protocol Failure}: The agent fails to issue the environment-specific termination command or prematurely terminates before completing the task.
\end{enumerate}

\begin{figure*}
    \includegraphics[width=\textwidth]{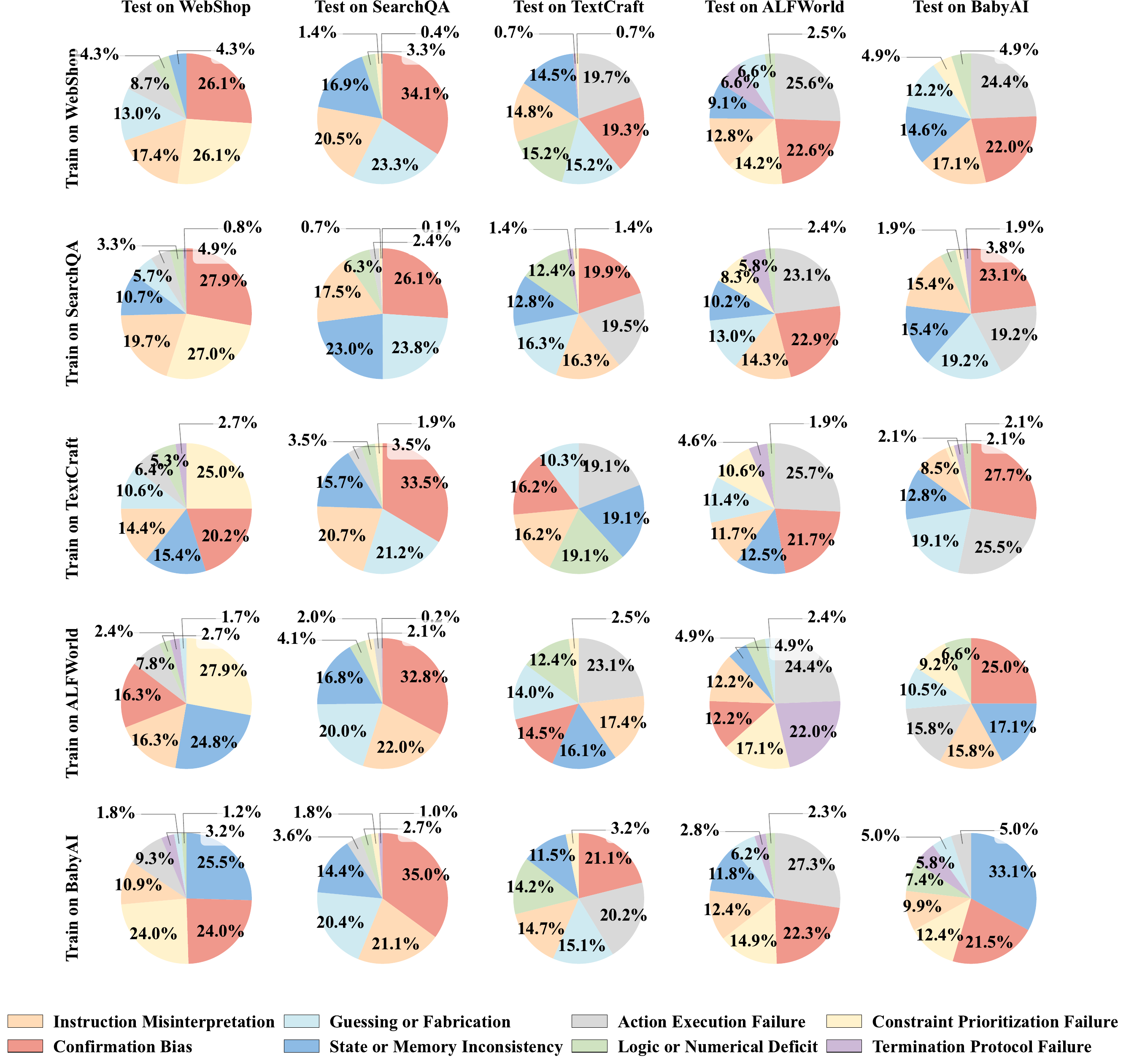}
    \caption{Failure mode distribution across all train-test environment combinations.}
    \label{fig:error_matrix}
\end{figure*}

The results of both in-domain and out-of-domain evaluations are discussed in Section \ref{sec:discussion}. Here, we present a detailed error classification in Figure \ref{fig:error_matrix}.

\section{Case Study}\label{appendix:case_study}

In this section, we present specific case studies. It is worth noting that while all interactions are inherently text-based, we provide visualizations for selected cases to enhance clarity. Additionally, due to the excessive number of interaction turns, we omit the majority of intermediate steps, highlighting only the pivotal moments in the figures.

\begin{figure*}[ht]
  \centering
  \includegraphics[width=\textwidth]{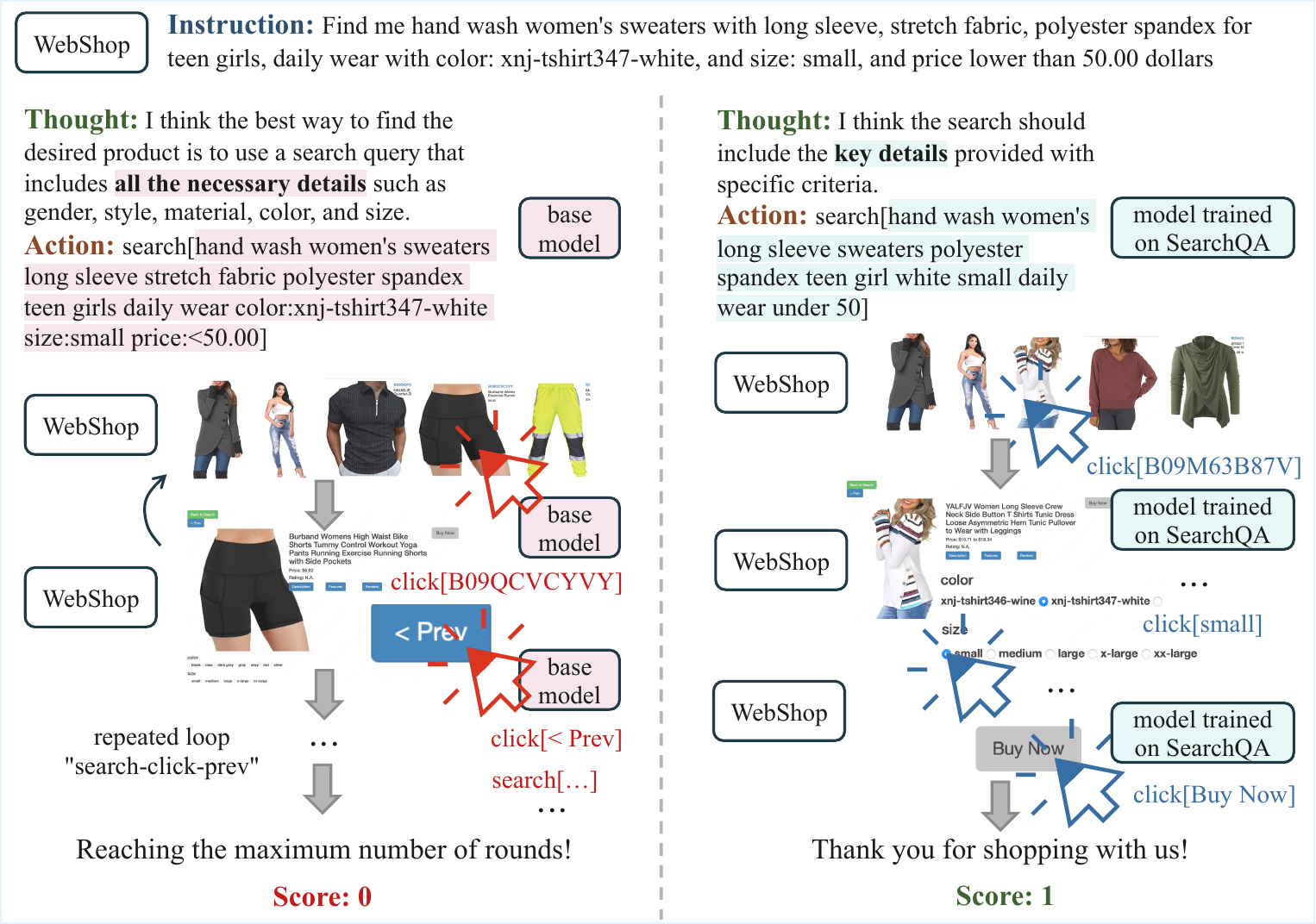}
  \caption{Comparison of trajectories on WebShop between the base model and the model trained on SearchQA.}
  \label{fig:case_webshop}
\end{figure*}

Figure \ref{fig:case_webshop} illustrates a case where a model trained on SearchQA is evaluated on WebShop, comparing its generated trajectory against that of the base model on the same task. As observed in the case, the base model tends to blindly input the entire content of the instruction into the search bar, resulting in the retrieval of numerous irrelevant items. Furthermore, the base model exhibits incoherent decision-making under multiple constraints and struggles to accurately extract key information from the voluminous HTML content returned by the environment, leading to the selection of items from incorrect categories. In contrast, the model trained on SearchQA learns to formulate more flexible search queries, as well as perform efficient information extraction from complex results, thereby enabling it to successfully retrieve key information and select the correct item.

\begin{figure*}[ht]
  \centering
  \includegraphics[width=\textwidth]{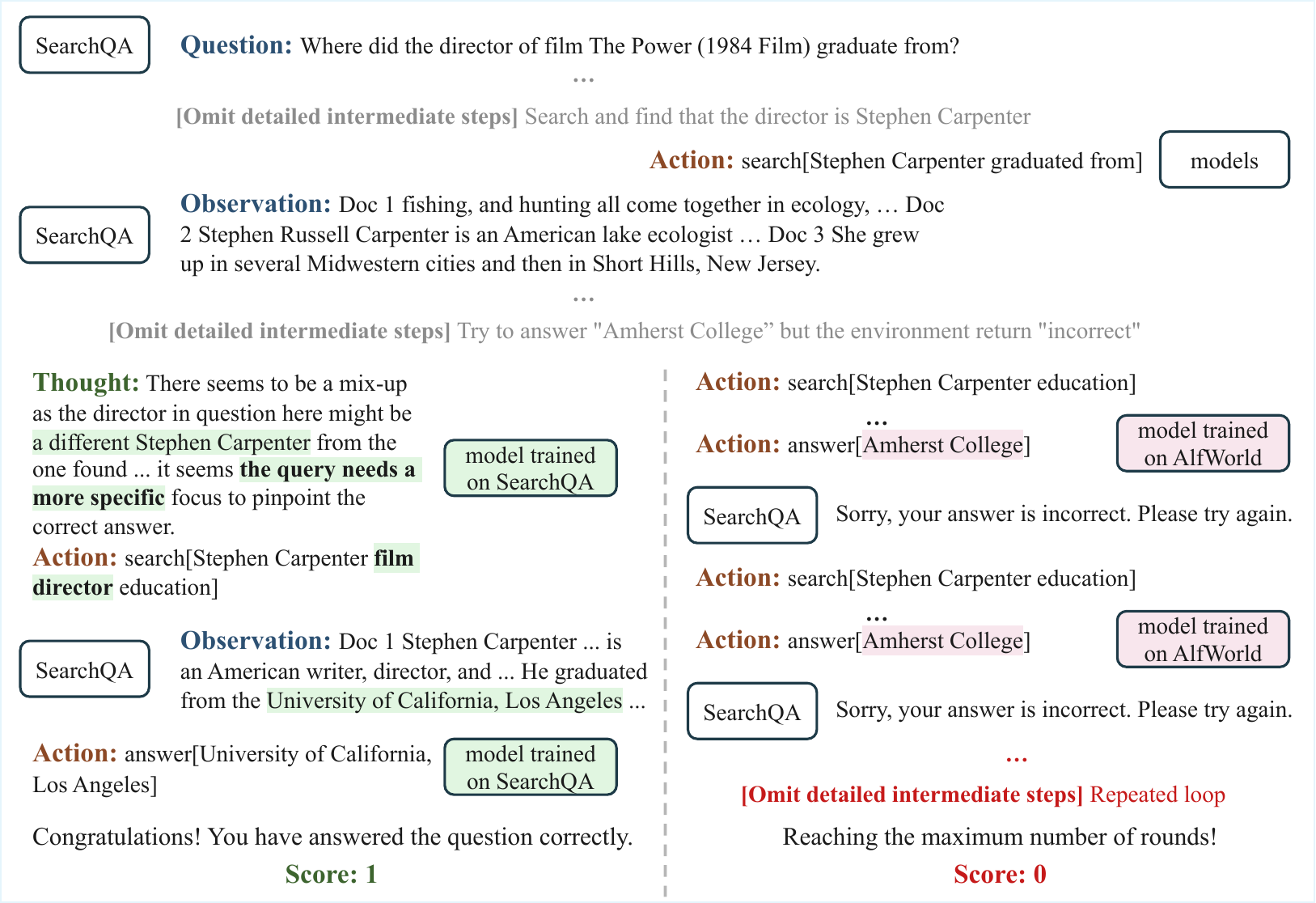}
  \caption{Comparison of trajectories on SearchQA between the model trained on SearchQA and the mode trained on AlfWorld.}
  \label{fig:case_searchqa}
\end{figure*}

Figure \ref{fig:case_searchqa} highlights a key reason why agents struggle to generalize to SearchQA, using model trained on AlfWorld as a case study. Although both models fail in their initial attempts, the model trained on SearchQA demonstrates the ability to refine its search queries to achieve greater precision. This capability proves difficult to transfer from other environments, e.g., the model trained on AlfWorld falls into a repetitive loop of the same searches and answers, consequently failing to resolve the task.

\begin{figure*}[ht]
  \centering
  \includegraphics[width=\textwidth]{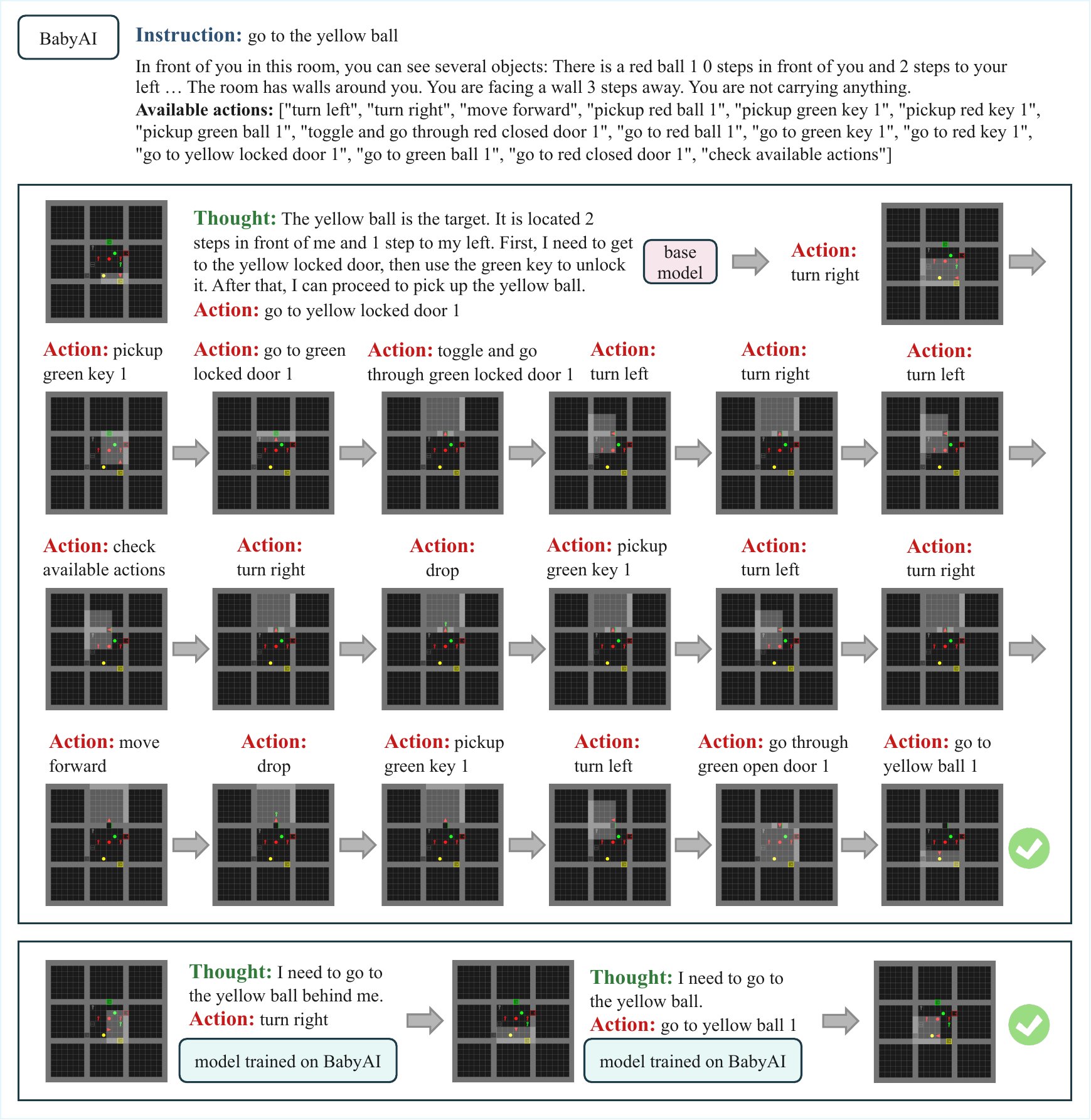}
  \caption{A case study in the BabyAI environment. Compared to the base model, the trained model demonstrates significantly improved exploration efficiency.}
  \label{fig:case_babyai}
\end{figure*}

Figure \ref{fig:case_babyai} illustrates how training enhances exploration efficiency, using BabyAI as a case study. The base model, lacking strong spatial awareness, struggles to accurately pinpoint the target location based on textual descriptions. Consequently, it falls into a pattern of redundant exploration; although it eventually completes the task, it requires an excessive number of interaction turns. In contrast, the trained model is able to precisely locate the target, thereby completing the task via an optimal path and significantly improving exploration efficiency.

\end{document}